\documentclass[%
 aip,
 amsmath,amssymb,
 reprint,%
]{revtex4-2}

\usepackage{graphicx}% Include figure files%图片和pdflatex一起用时的错误
\usepackage{dcolumn}% Align table columns on decimal point
\usepackage{bm}% bold math
\usepackage[utf8]{inputenc}%%%字体编码,用来嵌入字体
\usepackage[T1]{fontenc}%%%字体的编码
\usepackage{mathptmx}%%数学公式表达
\usepackage{etoolbox}%%表格的格式
\usepackage{siunitx} % 用于对齐小数点和短横线
\usepackage{empheq} %方程加括号
\usepackage{amsthm}  % 定理
\theoremstyle{plain}
\usepackage[margin=1in]{geometry}
\usepackage{times}
\usepackage{graphicx}
\usepackage{float}
\usepackage{booktabs}
\usepackage{amsmath}
\usepackage{tikz}
\usepackage{graphicx}
\usepackage{subfig}
\usepackage{overpic}%%%画图
\usepackage{xcolor} % 宏包用于颜色定义
\usepackage{caption} % 宏包用于自定义图片标题

\usepackage{mathptmx}
\DeclareMathAlphabet{\mathcal}{OMS}{cmsy}{m}{n}
\DeclareSymbolFont{largesymbols}{OMX}{cmex}{m}{n}

\usepackage[colorlinks,bookmarksopen,bookmarksnumbered,citecolor=blue!50!cyan,linkcolor=blue!50!cyan,urlcolor=blue!50!cyan, colorlinks=true]{hyperref}%参考文献加超链接
\usepackage{CJKutf8}
\usepackage{multirow}
\usepackage{booktabs}
\begin{document}

\preprint{AIP/123-QED}
%[Physics of Fluids]
\title{A point-based residual kolmogorov–arnold network for hemodynamic parameter prediction under variable inflow conditions}
% Force line breaks with \\School of Mathematical Sciences, Beijing Normal University, Xinjiekouwai Street, Beijing, 100875, China
\author{Jiaheng Kang (\begin{CJK*}{UTF8}{gbsn}康嘉恒
\end{CJK*})}
\affiliation{\mbox{School of Mathematical Sciences, Beijing Normal University, Xinjiekouwai Street, Beijing, 100875, China}}
\author{Jun Liu (\begin{CJK*}{UTF8}{gbsn}刘君
\end{CJK*})}
\author{Yuping Duan (\begin{CJK*}{UTF8}{gbsn}段玉萍
\end{CJK*})}
\email[E-mail addresses: ]{doveduan@gmail.com}
 
\affiliation{\mbox{School of Mathematical Sciences, Beijing Normal University, Xinjiekouwai Street, Beijing, 100875, China}}
%\affiliation{\mbox{Center for Applied Mathematics, Tianjin University, Weijin Road, Tianjin, 300072, China}}

%\\This line break forced with \textbackslash\textbackslash

%\renewcommand*{\thefootnote}{\alph{footnote}}
%日期%\date{\today}% It is always \today, today,
             %  but any date may be explicitly specified

\begin{abstract}
Intracranial aneurysm hemodynamics can be investigated using computational fluid dynamics (CFD), but repeated simulations across vascular geometries and inflow conditions are computationally expensive. We propose Point-ResKAN, a neural-operator-inspired point-cloud surrogate for predicting pressure and velocity fields in intracranial aneurysms. The architecture uses two PointNet-style encoders to represent vascular geometry and prescribed inflow boundary data separately. Its central component is a residual Kolmogorov--Arnold network (ResKAN) that combines nonlinear Jacobi-polynomial mappings with learnable affine residual branches, providing an additional feature-transformation path while accommodating changes in feature dimension. We further extend the framework to transient velocity-field prediction by conditioning velocity-increment estimates on the current velocity state, boundary information, and temporal encoding. On the Aneumo dataset, Point-ResKAN achieves lower pressure and velocity errors than the evaluated baselines on held-out geometries and selected unseen inflow conditions, with inference below one second after training. On a pulsatile, non-Newtonian aneurysm benchmark, it achieves relative \(L_2\) velocity errors of \(1.16\%\) for one-step prediction and \(6.89\%\) for autoregressive rollout, outperforming the compared models under both protocols. A supplementary Ahmed-body benchmark evaluates the ResKAN module for surface-pressure prediction. These results support Point-ResKAN as an efficient surrogate for steady and transient CFD-field approximation under the evaluated conditions.

%\noindent{\textbf{Keywords:}Computational Fluid Dynamics; PointNet; Physics-Informed Neural Networks; Quadratic Residual Net; 4D Hemodynamics}
\end{abstract}

\maketitle
\setlength{\parindent}{2em}

\section{Introduction}
Intracranial aneurysms (IAs) are localized dilations of cerebral arteries, with an estimated prevalence of \(3.6\%\)--\(6\%\) among adults older than 30 years.\cite{wardlaw2000detection,zakeri2024comprehensive,penn2011hemodynamic} Although most unruptured IAs remain asymptomatic, rupture may lead to aneurysmal subarachnoid hemorrhage, which is associated with mortality rates exceeding \(50\%\).\cite{habibi2024prediction} Hemodynamic factors are closely related to aneurysm initiation, growth, and rupture.\cite{jeong2012hemodynamics,detmer2020comparison, cuadrado2018cerebral,samady2011coronary} In particular, vascular geometry and inflow conditions jointly determine complex pressure and velocity patterns, including jet impingement, flow recirculation, and regions of elevated or oscillatory wall shear. Quantitative prediction of these hemodynamic parameters is therefore important for understanding aneurysm progression and rupture mechanisms.

Patient-specific hemodynamic assessment can be performed using invasive measurements, non-invasive imaging, or computational modeling.\cite{scheeren2019new} Advanced imaging techniques, such as four-dimensional flow magnetic resonance imaging, provide valuable flow information but remain limited by spatiotemporal resolution, acquisition time, hardware requirements, and operational cost.\cite{pour2019accuracy,markl20124d,kissas2020machine,rizk20214d} Computational fluid dynamics (CFD) has consequently become an important tool for resolving pressure, velocity, and wall-related flow quantities in patient-specific vascular geometries.\cite{jung2008three,behbahani2009review,campbell2012effect, lee2019identification,zhang2020numerical,antiga2008image} A typical CFD workflow requires anatomical reconstruction, mesh generation, boundary-condition specification, numerical solution of the Navier--Stokes equations, and post-processing.\cite{wang2013high} Although CFD provides detailed flow information, each geometry--boundary-condition configuration generally requires an independent simulation, resulting in substantial computational cost and workflow complexity.

Deep-learning-based surrogate models offer a possible route to rapid flow-field prediction after offline training.\cite{ribeiro2020deepcfd} Early studies applied convolutional and generative models to fluid-flow prediction under varying physical parameters.\cite{lee2017prediction} Related approaches have since been developed for cardiovascular applications, including hemodynamic prediction from geometric representations of carotid and other vascular domains.\cite{li2021prediction,arzani2022machine,wang2023deep, zhang2023physics,kang2024four} Recent studies have combined reduced-order representations with temporal learning for cardiovascular surrogate modeling. Barzegar Gerdroodbary and Salavatidezfouli developed a POD--LSTM surrogate for blood flow in carotid arteries with and without stenosis, examining prediction accuracy across cardiac phases and flow conditions.\cite{barzegar2025predictive} They also compared POD--LSTM and CNN--LSTM approaches for cardiovascular fluid--structure interaction in abdominal aortic aneurysms, evaluating predictions of hemodynamic and structural quantities.\cite{gerdroodbary2025predictive} Kheiri et al. investigated POD--Transformer and POD--LSTM models for cerebral aneurysm hemodynamics under parent-vessel dilation.\cite{kheiri2026cfd} These studies demonstrate the potential of reduced representations and temporal learning for efficient cardiovascular prediction. Complementing this line of research, Point-ResKAN combines geometry and inflow encoding with residual Jacobi-KAN mappings to predict fields on irregular vascular point clouds, with an additional extension to transient velocity prediction.

Physics-informed neural networks and neural operators provide two important alternatives for learning PDE solutions. Physics-informed neural networks incorporate governing equations and boundary conditions through residual-based losses.\cite{raissi2019physics,eivazi2022physics} They have been applied to fluid mechanics and biomedical modeling. \cite{sahli2020physics,mao2020physics,chen2022flowdnn, almajid2022prediction,buhendwa2021inferring} Nevertheless, they often require a new optimization procedure for each geometry or parameter configuration, which limits their efficiency for repeated prediction. Neural operators instead aim to learn mappings from problem inputs to solution fields across families of PDE instances. \cite{li2020fourier,anandkumar2020neural,he2023mgno} Once trained, they can evaluate new parameter configurations without solving a separate optimization problem for each case.

Applying operator-learning methods to patient-specific hemodynamics remains challenging because vascular domains are highly irregular and vary substantially across subjects. Some approaches map irregular domains to a common reference representation, as in geometry-aware Fourier and graph-based neural operators.\cite{he2024geom,li2023fourier,li2023geometry} Although effective, such mappings can be difficult to construct for heterogeneous vascular topologies and may introduce additional preprocessing. Point-cloud representations offer an alternative by describing each geometry directly through unordered spatial samples. PointNet, PointNet++, and related architectures provide permutation-invariant feature extraction from such data.\cite{qi2017pointnet,qi2017pointnet++,ma2022rethinking} Building on these ideas, point-cloud operator-learning approaches, including PCNO and LAON, have been developed for PDE problems on irregular domains.\cite{zeng2025point,sun2026latent} However, hemodynamic prediction requires the simultaneous representation of vascular geometry and inflow boundary data, whose coupled effects can produce strongly nonlinear changes in pressure and velocity.\cite{campbell2012effect,madhavan2018effect,conijn2021computational}

Kolmogorov--Arnold Networks (KANs) provide a flexible mechanism for nonlinear approximation by replacing fixed nodewise activation functions with learnable univariate mappings on network edges.\cite{liu2024kan} Recent KAN-based models have demonstrated potential for flow prediction on irregular domains, including KA-PointNet and physics-informed KAN formulations.\cite{kashefi2025kolmogorov,khedr2025physics} However, stacking polynomially parameterized KAN layers can lead to optimization difficulties and unstable feature propagation, particularly for large-scale point-cloud inputs. Moreover, a single geometric encoding branch does not explicitly distinguish vascular-domain information from the prescribed inflow condition.

To address these challenges, we propose Point-ResKAN, a neural-operator-inspired point-cloud surrogate for predicting pressure--velocity fields from vascular geometry and sampled inflow boundary data. Point-ResKAN employs two PointNet-style branches: a geometry branch extracts pointwise and global representations of the vascular domain, while an inflow branch encodes the prescribed inlet condition. The resulting features are fused at each point and processed by residual Jacobi-based KAN layers. Each ResKAN layer combines an edge-wise nonlinear Jacobi-polynomial mapping with a learnable affine residual branch, providing an additional feature-transformation path and accommodating different input and output feature dimensions. 

We further extend Point-ResKAN to transient velocity-field prediction and evaluate it using pulsatile, non-Newtonian CFD data over a cardiac cycle. The extended model predicts velocity increments conditioned on the current velocity state, boundary information, and temporal encoding. One-step prediction and autoregressive rollout assess both individual-step accuracy and error accumulation during successive predictions. This extension broadens the evaluation from steady Newtonian flow to time-dependent flow with shear-dependent viscosity.

The main contributions of this work are summarized as follows:
\begin{itemize}
        \item We develop a dual-branch point-cloud framework that separately encodes vascular geometry and sampled inflow boundary data and integrates these representations for pointwise pressure--velocity prediction on irregular vascular domains.

        \item We introduce ResKAN layers that combine nonlinear Jacobi-polynomial mappings with learnable affine residual branches, providing additive feature-transformation paths that accommodate changes in latent feature dimension.
        
        \item We extend Point-ResKAN to transient velocity-field prediction and evaluate the extension on pulsatile, non-Newtonian aneurysm CFD data over a cardiac cycle. Both one-step prediction and autoregressive rollout are examined to assess prediction accuracy and error accumulation.
        
        \item We evaluate steady pressure--velocity prediction on held-out geometries within the dataset distribution and selected extrapolated inflow rates, supported by architectural comparisons and sensitivity analyses. 
        %A supplementary Ahmed-body benchmark assesses ResKAN for surface-pressure prediction.
\end{itemize}

\begin{figure*}[t]
 \centering
 \includegraphics[width=0.785\linewidth]{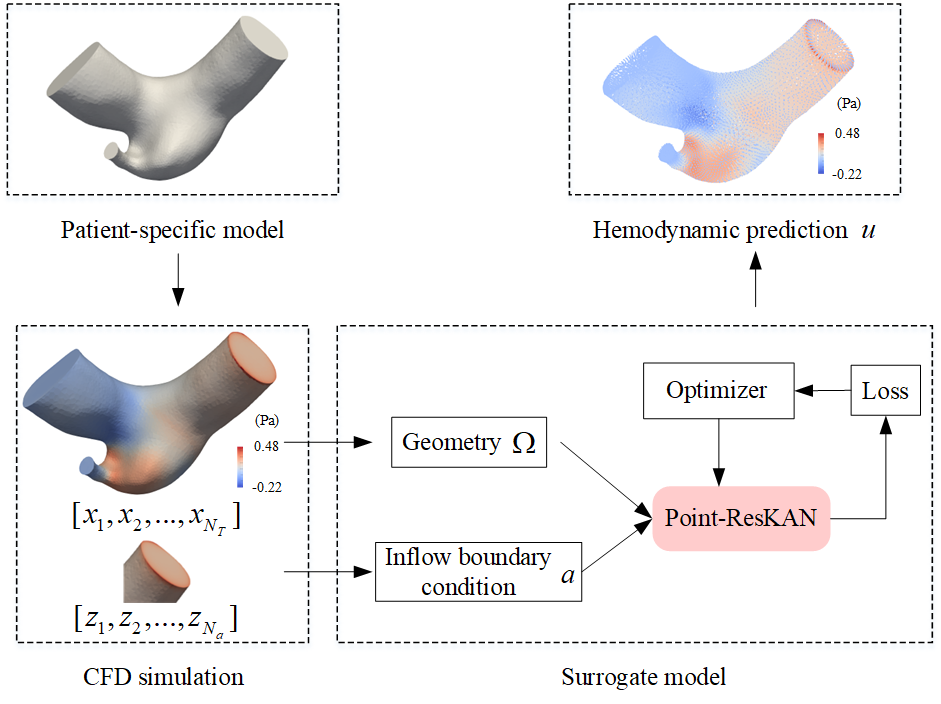}
 \caption{Optimization of the surrogate model for hemodynamic prediction.}
 \label{fig:1}
\end{figure*}

%%%%%%%%%%%%%%%%%%第二章节--方法学  \;  \emptyset \mathcal{U}

\section{Problem Formulation}
\label{sec:problem_formulation}

\subsection{Problem Setup and Parameter-to-Solution Map}
\label{sec:problem_setup}

We first formulate the steady pressure--velocity prediction task using the Aneumo dataset. Let \(D\subset\mathbb{R}^{3}\) be a fixed bounded reference domain containing the vascular geometries considered in this study. Each geometry is indexed by a signed distance function \(T\in\mathcal{T}\subset C^{0,1}(D)\), with $\Omega_T=\{x\in D:T(x)<0\}$ and $\partial\Omega_T=\{x\in D:T(x)=0\}$. 
The admissible class \(\mathcal{T}\) is restricted to geometries for which \(\Omega_T\) is a bounded Lipschitz domain contained in the interior of \(D\). We use the convention that \(T\) is negative inside the vessel and positive outside. The signed-distance representation is used here to index geometries in the continuous problem formulation. However, the implemented network receives sampled spatial coordinates.

Let \(\mathcal{A}\subset\mathbb{R}^{d_a}\) denote the admissible inflow-parameter space. For each geometry, the boundary is decomposed as
$\partial\Omega_T
=
\overline{\Gamma}_{\mathrm{in}}
\cup
\overline{\Gamma}_{\mathrm{out}}
\cup
\overline{\Gamma}_{\mathrm{wall}}$, 
where the inlet, outlet, and wall are pairwise disjoint, relatively open boundary subsets whose closures may intersect along their interfaces. Their dependence on \(T\) is suppressed for simplicity. For the steady-flow Aneumo benchmark, blood is represented as an incompressible, homogeneous Newtonian continuum. The homogeneous-continuum approximation does not explicitly resolve cellular constituents or spatial variations in hematocrit, while the Newtonian approximation assumes constant viscosity. Under the additional steady-flow assumption, the velocity \(\mathbf{u}:\Omega_T\rightarrow\mathbb{R}^{3}\) and pressure \(P:\Omega_T\rightarrow\mathbb{R}\) satisfy
\begin{subequations}\label{eq:navier_stokes}
\begin{empheq}[left=\empheqlbrace]{align}
\nabla\cdot\mathbf{u}&=0,&&\text{in }\Omega_T,\\
\rho(\mathbf{u}\cdot\nabla)\mathbf{u}&=-\nabla P+\mu\nabla^2\mathbf{u}+\rho\mathbf{b},&&\text{in }\Omega_T,\\
\mathbf{u}&=\mathbf{0},&&\text{on }\Gamma_{\mathrm{wall}},\\
\mathbf{u}&=\mathbf{g}_{\mathrm{in}}(\cdot;a),&&\text{on }\Gamma_{\mathrm{in}},\\
\mathcal{B}_{\mathrm{out}}(\mathbf{u},P)&=q_{\mathrm{out}},
&&\text{on }\Gamma_{\mathrm{out}},
\end{empheq}
\end{subequations}
where \(\rho\) and \(\mu\) are the blood density and dynamic viscosity, respectively;\cite{ab2016prediction}
\(\mathbf{b}\) denotes body-force acceleration, which may be zero; and
\(\mathbf{g}_{\mathrm{in}}(\cdot;a)\) is the prescribed inlet velocity profile parameterized by \(a\in\mathcal{A}\). The operator \(\mathcal{B}_{\mathrm{out}}\) represents the outlet condition of the reference CFD problem. For the Aneumo simulations, a zero-gauge-pressure condition is prescribed at the outlet, fixing the pressure reference. The vessel walls are stationary and rigid, with no slip.

For each admissible pair \((a,T)\), let
$\boldsymbol{y}_{a,T}
=
(P_{a,T},\mathbf{u}_{a,T})
\in\mathcal{U}(\Omega_T)$ 
denote the corresponding reference solution. For the purpose of defining the target map, we assume that the reference problem selects a unique solution within the geometries and inflow conditions considered. To express solutions on a common domain, let $\operatorname{Ext}_T:
\mathcal{U}(\Omega_T)
\longrightarrow
L^p(D;\mathbb{R}^{d+1})$, $1\leq p<\infty$,
denote a fixed extension convention, such as extension by zero outside \(\Omega_T\). The continuous parameter-to-solution map is
\begin{equation}
\begin{aligned}
G^\dagger: &\mathcal{A}\times\mathcal{T}\longrightarrow L^p(D;\mathbb{R}^{d+1}),\\
&G^\dagger(a,T)=\operatorname{Ext}_T\boldsymbol{y}_{a,T}.
\end{aligned}
\label{eq:parameter_to_solution}
\end{equation}
The extension introduces a common continuous codomain for solutions defined on different vascular domains and does not imply that the surrogate operates on a fixed spatial discretization. The continuous map defines the physical approximation target and does not establish discretization independence of the learned surrogate.

At the discrete level, the vascular domain is represented by the indexed point cloud $\mathcal{X}_T=\{x_i\}_{i=1}^{N_T}$, $x_i=(x_{i, 1},\ldots,x_{i, d}) \in\overline{\Omega}_T$. The corresponding inflow boundary data are represented by $\mathcal{Z}_{a,T}=\{z_j\}_{j=1}^{N_a}$, $z_j=\left( \xi_j,\mathbf p_{\mathrm{in}}(\xi_j) , \mathbf{g}_{\mathrm{in}}(\xi_j;a) \right) \in \mathbb{R}^{2d+1}$, $\xi_j\in\Gamma_{\mathrm{in}}$. The dependence of \(\mathcal{Z}_{a, T}\) on \(T\) reflects the geometry-dependent inlet locations. 
The indexed point clouds \(\mathcal{X}_T\) and \(\mathcal{Z}_{a, T}\) constitute the actual inputs to Point-ResKAN, whose target output is $\left\{\boldsymbol{y}_{a, T}(x_i)\right\}_{i=1}^{N_T}$. The overall workflow for predicting the hemodynamic parameters is shown in Fig.~\ref{fig:1}.

\subsection{Neural-Operator-Inspired Surrogate Framework}
\label{sec:surrogate_framework}

Many neural-operator architectures follow a lifting--latent-transformation--projection structure, which motivates the organization of our surrogate:

\begin{equation}
\mathcal{N}_{\theta}=\mathcal{R} \circ \mathcal{V}_{L} \circ \cdots \circ \mathcal{V}_{1} \circ \mathcal{L},
\label{eq:generic_operator}
\end{equation}
where \(\mathcal{L}\) lifts the inputs into a latent feature space, \(\{\mathcal{V}_{l}\}_{l=1}^{L}\) transform the latent features, and \(\mathcal{R}\) maps the resulting representation to the output variables. The parameter set \(\theta\) contains all trainable weights.

Following this organization, Point-ResKAN learns a shared mapping from vascular geometry and sampled inlet data to pressure--velocity fields across multiple steady-flow cases. It operates on finite point clouds using separate geometry and inflow encoders, pointwise feature fusion, and a shared decoder. Its mapping is written as
\begin{equation}
\mathcal{S}_{\theta}(\mathcal{X}_T,\mathcal{Z}_{a,T})
=
\mathcal{R}\!\left(
\mathcal{F}_{\mathrm{fuse}}\!\left(
\mathcal{E}_{\mathrm{geo}}(\mathcal{X}_T),
\mathcal{E}_{\mathrm{in}}(\mathcal{Z}_{a,T})
\right)\right),
\label{eq:point_reskan_operator}
\end{equation}
where \(\mathcal{E}_{\mathrm{geo}}\) and \(\mathcal{E}_{\mathrm{in}}\) encode the vascular geometry and inlet data, respectively. The ResKAN transformations are embedded within these encoder branches. The fusion operation \(\mathcal{F}_{\mathrm{fuse}}\) combines the encoded features at each vascular point, and \(\mathcal{R}\) decodes them into pressure and velocity.

For a vascular point cloud
\(\mathcal{X}_T=\{x_i\}_{i=1}^{N_T}\subset\overline{\Omega}_T\)
and an inlet point cloud \(\mathcal{Z}_{a,T}\), the encoder outputs are
\begin{equation}
\{h_i^{\mathrm{geo}}\}_{i=1}^{N_T}
=\mathcal{E}_{\mathrm{geo}}(\mathcal{X}_T),\quad
h_{a,T}^{\mathrm{in}}
=\mathcal{E}_{\mathrm{in}}(\mathcal{Z}_{a,T}),
\label{eq:surrogate_encoder_outputs}
\end{equation}
where \(h_i^{\mathrm{geo}}\) combines pointwise geometric features with global vascular context, and \(h_{a,T}^{\mathrm{in}}\) is a global representation of the sampled inlet data. The inlet representation is shared across vascular points during fusion:
\begin{equation}
v_i^{\mathrm{fuse}}
=\mathcal{F}_{\mathrm{fuse}}
\left(h_i^{\mathrm{geo}},h_{a,T}^{\mathrm{in}}\right),\quad
\widehat{\boldsymbol{y}}_{a,T}(x_i)
=\mathcal{R}\left(v_i^{\mathrm{fuse}}\right),
\label{eq:surrogate_fusion_output}
\end{equation}
with
\(\widehat{\boldsymbol{y}}_{a,T}(x_i)
=(\widehat{P}_{a,T}(x_i),\widehat{\mathbf{u}}_{a,T}(x_i))\).
By collecting the pointwise predictions, it gives
\begin{equation}
\mathcal{S}_{\theta}
\left(\mathcal{X}_T,\mathcal{Z}_{a,T}\right)
=
\left\{
\widehat{\boldsymbol{y}}_{a,T}(x_i)
\right\}_{i=1}^{N_T}.
\label{eq:point_surrogate}
\end{equation}
The neural-operator-inspired design learns a shared mapping from vascular geometry and inlet data to solution fields through an encoding--latent-transformation--decoding architecture. Symmetric aggregation yields permutation-invariant global features, while shared pointwise operations produce outputs that are equivariant to permutations of the vascular points. The architecture accommodates variable point counts, but these properties do not establish prediction consistency under resampling or changes in discretization. Point-ResKAN is therefore evaluated as a point-cloud CFD surrogate approximating the continuous solution map defined in Section~\ref{sec:problem_setup}.

\begin{figure*}
\centering
\includegraphics[width=0.9\linewidth]{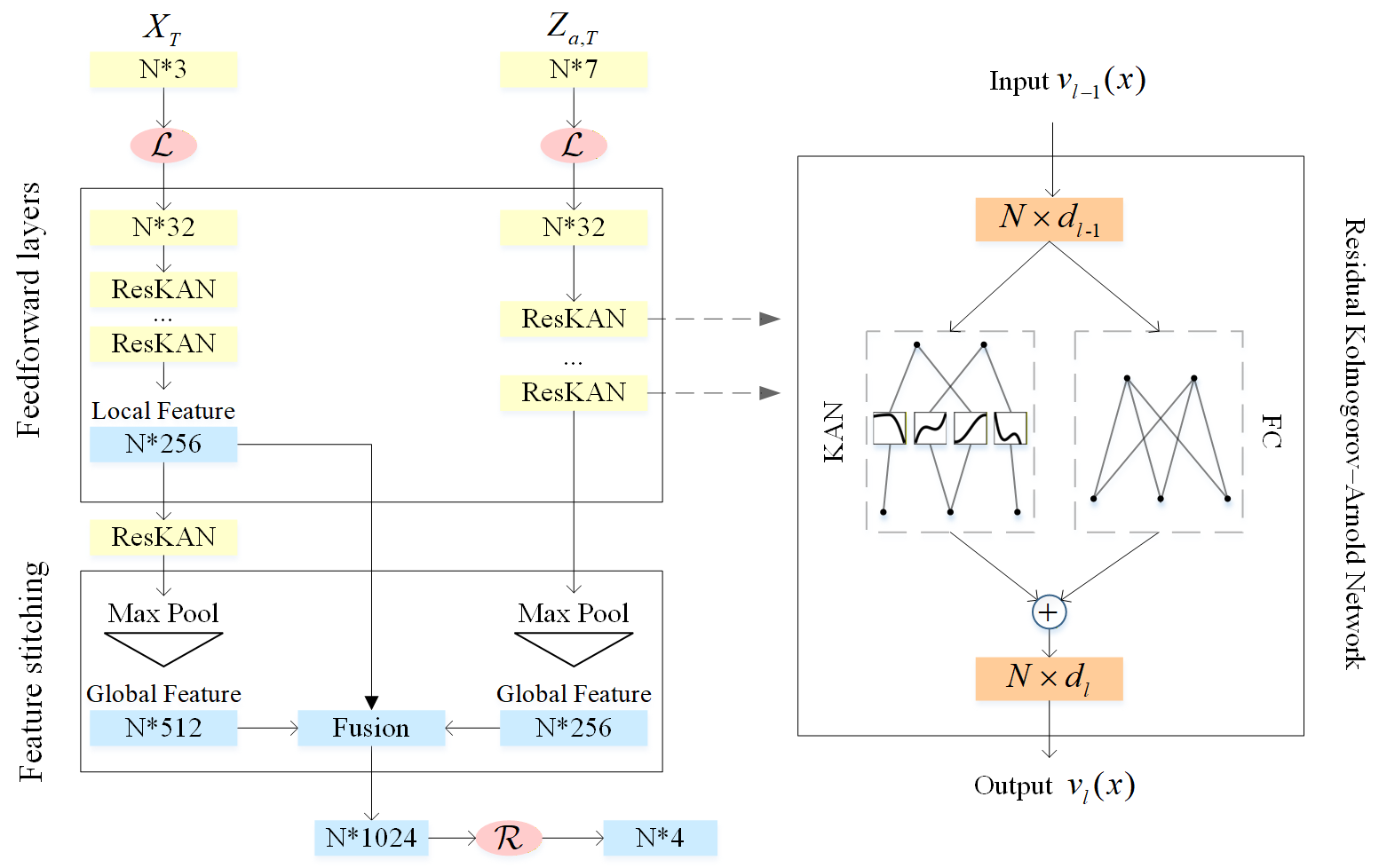}
\caption{Architecture of Point-ResKAN. The geometry and inflow encoders use lifting, ResKAN transformations, and feature aggregation to extract local--global vascular features and a global inflow representation, respectively. These features are fused pointwise and projected by \(\mathcal{R}\) onto the pressure--velocity field.}
\label{fig:2}
\end{figure*}

\subsection{Nonlinear Latent Mapping via ResKAN}
\label{sec:reskan_mapping}

As described in Sec.~\ref{sec:surrogate_framework}, the geometry and inflow encoders combine input lifting, ResKAN transformations, and symmetric feature aggregation. Their global feature-extraction paths follow
$g\circ\mathcal{V}_{L}\circ\cdots\circ\mathcal{V}_{1}
\circ\mathcal{L}$, where \(\mathcal{L}\) is the lifting map,
\(\{\mathcal{V}_{l}\}_{l=1}^{L}\) are ResKAN transformations,
and \(g\) denotes max pooling over the input points. The two encoders use independent trainable parameters. The geometry encoder combines its pooled global feature with pointwise features to construct \(h_i^{\mathrm{geo}}\), whereas the inflow encoder produces the global representation \(h_{a,T}^{\mathrm{in}}\).

For a pointwise input feature
\(\mathbf{v}_{l-1}\in\mathbb{R}^{d_{l-1}}\),
the \(l\)-th ResKAN layer first applies a componentwise
hyperbolic tangent:
\begin{equation}
\bar{\mathbf{v}}_{l-1}
=
\tanh(\mathbf{v}_{l-1})
\in(-1,1)^{d_{l-1}}.
\label{eq:kan_normalization}
\end{equation}
The Jacobi-based KAN mapping
\(\Phi_l:(-1,1)^{d_{l-1}}\rightarrow\mathbb{R}^{d_l}\)
is defined by
\begin{equation}
\left[\Phi_l(\bar{\mathbf{v}}_{l-1})\right]_j
=
\sum_{k=1}^{d_{l-1}}
\phi_{j,k}^{(l)}
\left([\bar{\mathbf{v}}_{l-1}]_k\right),
\quad j=1,\ldots,d_l,
\label{eq:kan_layer}
\end{equation}
where each edge function \(\phi_{j,k}^{(l)}\) is represented
by a truncated Jacobi-polynomial expansion:
\begin{equation}
\phi_{j,k}^{(l)}(\zeta)
=
\sum_{m=0}^{n}
c_{m,j,k}^{(l)}
\mathsf{J}_{m}^{(\alpha,\beta)}(\zeta),
\qquad \zeta\in[-1,1].
\label{eq:jacobi_expansion}
\end{equation}
Here, \(n\) is the maximum polynomial degree,
\(c_{m,j,k}^{(l)}\) are trainable coefficients, and
\(\mathsf{J}_{m}^{(\alpha,\beta)}\) is the degree-\(m\)
Jacobi polynomial with fixed parameters \(\alpha,\beta>-1\).

The polynomials are evaluated using
\[
\mathsf{J}_{0}^{(\alpha,\beta)}(\zeta)=1,
\quad
\mathsf{J}_{1}^{(\alpha,\beta)}(\zeta)
=
\frac{\alpha+\beta+2}{2}\zeta
+
\frac{\alpha-\beta}{2},
\]
and the three-term recurrence
\begin{equation}
\mathsf{J}_{m}^{(\alpha,\beta)}(\zeta)
=
(A_m\zeta+B_m)
\mathsf{J}_{m-1}^{(\alpha,\beta)}(\zeta)
+
C_m\mathsf{J}_{m-2}^{(\alpha,\beta)}(\zeta),
~~ m\geq2,
\label{eq:jacobi_recurrence}
\end{equation}
where
\[
\begin{aligned}
A_m
&=
\frac{(2m+\alpha+\beta-1)(2m+\alpha+\beta)}
{2m(m+\alpha+\beta)},\\
B_m
&=
\frac{(2m+\alpha+\beta-1)(\alpha^2-\beta^2)}
{2m(m+\alpha+\beta)(2m+\alpha+\beta-2)},\\
C_m
&=
-\frac{(m+\alpha-1)(m+\beta-1)(2m+\alpha+\beta)}
{m(m+\alpha+\beta)(2m+\alpha+\beta-2)}.
\end{aligned}
\]

To complement the nonlinear mapping, we introduce a deterministic affine residual branch:
\begin{equation}
\boldsymbol{\delta}_l(\mathbf{v}_{l-1})
=
W_l\mathbf{v}_{l-1}+\mathbf{b}_l,
\label{eq:affine_residual}
\end{equation}
where \(W_l\in\mathbb{R}^{d_l\times d_{l-1}}\) and \(\mathbf{b}_l\in\mathbb{R}^{d_l}\) are trainable parameters. The complete ResKAN transformation is
\begin{equation}
\begin{aligned}
\mathbf{v}_l
&=\mathcal{V}_l(\mathbf{v}_{l-1})\\
&=
\Phi_l\!\left(\tanh(\mathbf{v}_{l-1})\right)
+
W_l\mathbf{v}_{l-1}
+
\mathbf{b}_l,
\qquad l=1,\ldots,L.
\end{aligned}
\label{eq:reskan_layer}
\end{equation}
The affine branch acts on the input features before the \(\tanh\) transformation and accommodates unequal input and output dimensions. The residual connection is an additive feature-space operation, rather than a residual of the governing equations.

The Jacobian of the layer with respect to its input can be obtained by
\begin{equation}
\frac{\partial\mathbf{v}_l}
{\partial\mathbf{v}_{l-1}}
=
\mathrm{D}\Phi_l\!\left(\tanh(\mathbf{v}_{l-1})\right)
\operatorname{Diag}\!\left(
1-\tanh^2(\mathbf{v}_{l-1})
\right)
+
W_l,
\label{eq:reskan_gradient}
\end{equation}
where \(\mathrm{D}\Phi_l\) denotes the Jacobian of the
Jacobi-based KAN mapping. The additive term \(W_l\)
provides a gradient path that does not pass through
the derivative of \(\tanh\). This structural property
motivates the affine branch but does not, by itself,
guarantee stable gradients or improved convergence. 
Stacking the layers gives
\begin{equation}
\mathbf{v}_L
=
\left(
\mathcal{V}_{L}\circ\cdots\circ\mathcal{V}_{1}
\right)(\mathbf{v}_0),
\label{eq:stacked_reskan}
\end{equation}
which defines the latent feature transformation within
each encoder. The resulting pointwise and pooled features
form the geometry and inflow representations used in
Sec.~\ref{sec:surrogate_framework}.

\subsection{Geometry and Inflow Encoders}
\label{sec:point_encoders}

The geometry and inflow encoders apply shared pointwise ResKAN transformations within each branch, followed by componentwise max pooling. The two branches have independent trainable parameters. For the vascular point cloud
\(\mathcal{X}_T=\{x_i\}_{i=1}^{N_T}\),
the geometry encoder computes
\[
\mathbf{v}_i^{\mathrm{geo}}
=
\left(
\mathcal{V}_{L_{\mathrm{geo}}}^{\mathrm{geo}}
\circ\cdots\circ
\mathcal{V}_{1}^{\mathrm{geo}}
\circ\mathcal{L}_{\mathrm{geo}}
\right)(x_i),
\quad i=1,\ldots,N_T.
\]
Componentwise max pooling produces the global vascular feature
\[
\mathbf{v}_T^{\mathrm{global}}
=
g\!\left(
\{\mathbf{v}_i^{\mathrm{geo}}\}_{i=1}^{N_T}
\right).
\]
This feature is concatenated with each pointwise representation to provide global geometric context:
\[
\mathbf{h}_i^{\mathrm{geo}}
=
\left[
\mathbf{v}_i^{\mathrm{geo}}
\,\Vert\,
\mathbf{v}_T^{\mathrm{global}}
\right],
\qquad
\mathcal{E}_{\mathrm{geo}}(\mathcal{X}_T)
=
\{\mathbf{h}_i^{\mathrm{geo}}\}_{i=1}^{N_T},
\]
where \(\Vert\) denotes feature concatenation. For the sampled inlet data
\(\mathcal{Z}_{a,T}=\{z_j\}_{j=1}^{N_a}\),
the inflow encoder computes
\[
\mathbf{v}_j^{\mathrm{in}}
=
\left(
\mathcal{V}_{L_{\mathrm{in}}}^{\mathrm{in}}
\circ\cdots\circ
\mathcal{V}_{1}^{\mathrm{in}}
\circ\mathcal{L}_{\mathrm{in}}
\right)(z_j),
\qquad j=1,\ldots,N_a,
\]
and aggregates these features into a global inlet representation:
\[
\mathbf{h}_{a,T}^{\mathrm{in}}
=
\mathcal{E}_{\mathrm{in}}(\mathcal{Z}_{a,T})
=
g\!\left(
\{\mathbf{v}_j^{\mathrm{in}}\}_{j=1}^{N_a}
\right).
\]
The geometry encoder therefore combines pointwise features with global vascular context, while the inflow encoder summarizes the sampled inlet data. The inlet representation is broadcast to all vascular points and combined with \(\mathbf{h}_i^{\mathrm{geo}}\) through \(\mathcal{F}_{\mathrm{fuse}}\), as described in Sec.~\ref{sec:surrogate_framework}.

\subsection{Training Objective}
\label{sec:point_reskan_training}

The complete Point-ResKAN architecture is illustrated in Fig.~\ref{fig:2}. For the steady-flow task, let
\(\{\mathcal{X}_r\}_{r=1}^{N_{\mathrm{geo}}}\)
denote the training vascular point clouds, where
\(\mathcal{X}_r=\{x_{r,i}\}_{i=1}^{N_r}
\subset\overline{\Omega}_r\).
Let \(\{a_s\}_{s=1}^{N_{\mathrm{BC}}}\subset\mathcal{A}\)
denote the training inflow conditions, and let
\(\mathcal{Z}_{r,s}\) contain the sampled inlet data for
geometry \(r\) and condition \(a_s\).

For each geometry--inflow pair, the CFD reference field and
the corresponding prediction are
\[
\boldsymbol{Y}_{r,s}^{\mathrm{CFD}}
=
\{\boldsymbol{y}_{r,s}(x_{r,i})\}_{i=1}^{N_r},
\qquad
\widehat{\boldsymbol{Y}}_{r,s}
=
\mathcal{S}_{\theta}
(\mathcal{X}_r,\mathcal{Z}_{r,s}).
\]
Assuming that each training geometry is paired with all
\(N_{\mathrm{BC}}\) selected inflow conditions, the training
objective is
\begin{equation}
\theta^\ast
=
\arg\min_{\theta}
\frac{1}{N_{\mathrm{geo}}N_{\mathrm{BC}}}
\sum_{r=1}^{N_{\mathrm{geo}}}
\sum_{s=1}^{N_{\mathrm{BC}}}
\ell\left(
\widehat{\boldsymbol{Y}}_{r,s},
\boldsymbol{Y}_{r,s}^{\mathrm{CFD}}
\right),
\label{eq:training_objective}
\end{equation}
where \(\ell\) measures the discrepancy between the predicted
and reference fields on the sampled training points.
The specific loss formulation and normalization procedures
are detailed in Section IV.A. All trainable parameters of
\(\mathcal{S}_{\theta}\) are optimized jointly.

\section{Transient Velocity-Field Prediction}
\label{sec:transient_prediction}

We extend Point-ResKAN from steady pressure--velocity prediction to transient velocity-field prediction and evaluate the extension using pulsatile, non-Newtonian CFD data from the BenchAnXplore dataset.\cite{Lannelongue2026BenchAnXplore} The dataset authors generated the reference solutions using an in-house C++ finite-element solver with a variational multiscale formulation. The reference problem is governed by the unsteady incompressible Navier--Stokes equations:
\begin{subequations}
\label{eq:transient-ns}
\begin{empheq}[left=\empheqlbrace]{align}
\nabla\cdot\mathbf{u}
&=0,\\
\rho\left(
\partial_t\mathbf{u}
+
(\mathbf{u}\cdot\nabla)\mathbf{u}
\right)
&=
-\nabla P
+
\nabla\cdot
\left(
2\mu(\dot{\gamma})\boldsymbol{\epsilon}(\mathbf{u})
\right),
\end{empheq}
\end{subequations}
where
\[
\boldsymbol{\epsilon}(\mathbf{u})
=
\frac{1}{2}
\left(
\nabla\mathbf{u}+\nabla\mathbf{u}^{\mathsf T}
\right) ~~\mbox{and}~~
\dot{\gamma}
=
\sqrt{
2\boldsymbol{\epsilon}(\mathbf{u})
:
\boldsymbol{\epsilon}(\mathbf{u})
}
\]
are the strain-rate tensor and scalar shear rate, respectively.
The shear-dependent viscosity is described by the
Carreau--Yasuda model:
\begin{equation*}
\mu(\dot{\gamma})
=
\mu_\infty
+
(\mu_0-\mu_\infty)
\left[
1+(\tau\dot{\gamma})^a
\right]^{(n-1)/a},
\label{eq:carreau_yasuda}
\end{equation*}
where \(\mu_0\) and \(\mu_\infty\) are the limiting viscosities at zero and infinite shear rates, \(\tau\) is a characteristic time scale, and \(n\) and \(a\) control the shear-thinning behavior and transition between viscosity regimes.\cite{lynch2022effects} These rheological parameters are prescribed in the reference CFD model rather than learned by the surrogate.

\begin{figure*}[htbp]
 \centering
 \includegraphics[width=0.9\linewidth]{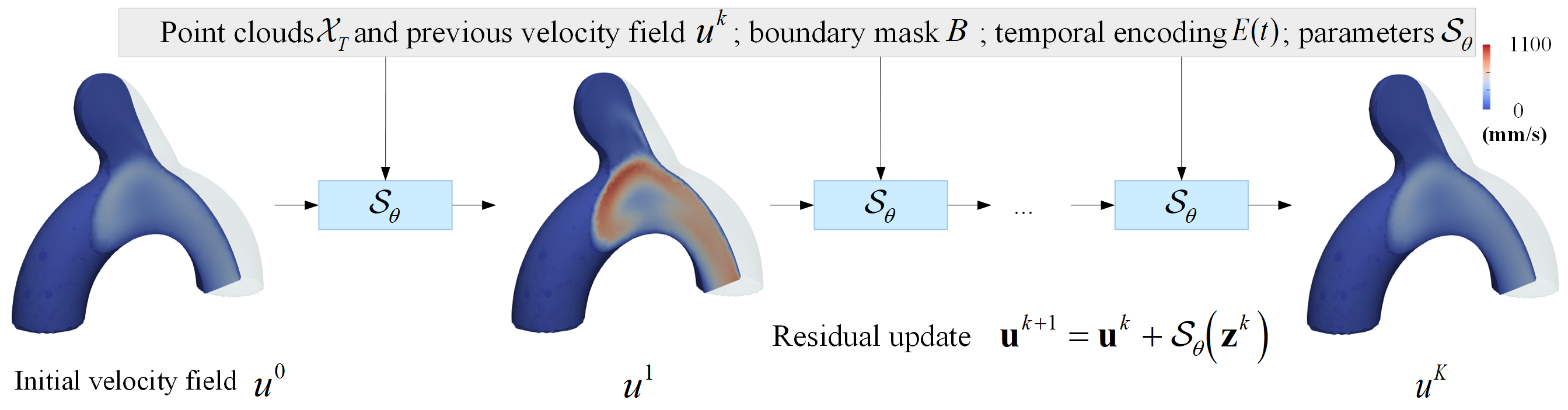}
 \caption{ Transient velocity-prediction workflow.
 The model predicts a velocity increment from spatial
 coordinates, the current velocity state, a wall-boundary
 mask, and temporal encoding. The increment is added to
 the current state, and zero velocity is imposed at wall
 nodes.}
 \label{fig:tr_model}
\end{figure*}

\textbf{Velocity-increment prediction.} The transient model learns a discrete state-update mapping at the temporal sampling interval of the dataset. For a fixed vascular point cloud \(\mathcal{X}_T=\{\mathbf{x}_i\}_{i=1}^{N_T}\), the input at time \(t^k\) is
\[
\mathbf{z}_i^k
=
\left[
\mathbf{x}_i,\,
\mathbf{u}_i^k,\,
B_i,\,
E(t^k)
\right],
\]
where \(\mathbf{u}_i^k\) is the current velocity and \(B_i\) is a binary mask equal to zero at wall nodes and one at the remaining nodes. The temporal encoding \(E(t^k)\) is shared by all points at the same time.

Let \(\mathcal{S}_{\theta}^{\mathrm{tr}}\) denote the transient network. Its output is a velocity increment:
\begin{equation}
\Delta\widehat{\mathbf{u}}_i^k
=
\left[
\mathcal{S}_{\theta}^{\mathrm{tr}}
\left(\{\mathbf{z}_j^k\}_{j=1}^{N_T}\right)
\right]_i,~~
\widehat{\mathbf{u}}_i^{k+1}
=
B_i\left(
\mathbf{u}_i^k+
\Delta\widehat{\mathbf{u}}_i^k
\right).
\label{eq:transient_update}
\end{equation}
The mask enforces zero velocity at the stationary vessel walls after each update. This explicit wall treatment does not impose the governing equations on the predicted interior field. The temporal encoding is
\begin{equation}
\begin{aligned}
E(t^k)
&=
\left[
\sin(\omega_j t^k),\,
\cos(\omega_j t^k)
\right]_{j=1}^{d_t/2},\\
\omega_j
&=
\exp\left(
-\frac{j-1}{d_t/2-1}\ln(10^4)
\right),
\qquad d_t=18.
\end{aligned}
\label{eq:time}
\end{equation}
The prediction workflow is illustrated in
Fig.~\ref{fig:tr_model}.

\textbf{Training objective.}
The transient model is trained on one-step transitions using the current CFD velocity as input and the next CFD velocity as the target. The normalized mean squared error excludes the wall nodes:
\begin{equation}
\ell_{\mathrm{tr}}
=
\frac{1}{3N_{\mathrm{valid}}}
\sum_{i\in\mathcal{I}_{\mathrm{valid}}}
\left\|
\frac{
\widehat{\mathbf{u}}_i^{k+1}
-
\mathbf{u}_i^{k+1}
}{s_u}
\right\|_2^2,
\label{eq:transient_loss}
\end{equation}
where \(\mathcal{I}_{\mathrm{valid}}=\{i:B_i=1\}\), \(N_{\mathrm{valid}}=|\mathcal{I}_{\mathrm{valid}}|\), and \(s_u\) is a fixed RMS velocity scale computed from the training data, excluding wall nodes. The factor of three accounts for the three velocity components. The same training-derived scale is used for the held-out data.

\textbf{One-step and autoregressive rollout evaluation.} In one-step evaluation, the current CFD velocity \(\mathbf{u}^k\) is supplied to the network, and \(\widehat{\mathbf{u}}^{k+1}\) is compared with the next CFD reference field. In autoregressive rollout, the model is initialized with a CFD velocity field and subsequently uses its own predictions:
\begin{equation}
\begin{aligned}
\widehat{\mathbf{u}}_i^0
&=\mathbf{u}_i^0,\quad
\widehat{\mathbf{z}}_i^k
=
\left[
\mathbf{x}_i,\,
\widehat{\mathbf{u}}_i^k,\,
B_i,\,
E(t^k)
\right],\\
\widehat{\mathbf{u}}_i^{k+1}
&=
B_i\left(
\widehat{\mathbf{u}}_i^k
+
\left[
\mathcal{S}_{\theta}^{\mathrm{tr}}
\left(
\{\widehat{\mathbf{z}}_j^k\}_{j=1}^{N_T}
\right)
\right]_i
\right).
\end{aligned}
\label{eq:transient_rollout}
\end{equation}
The two protocols assess one-step accuracy and error accumulation during successive predictions, respectively. Prediction errors are quantified using RMSE and relative \(L_2\) error. Training uses CFD-based one-step transitions; autoregressive rollout is used for evaluation.

\section{Results and Discussion}
\label{sec:results}

\subsection{Datasets and Numerical Implementation}
\label{sec:datasets_implementation}

\textbf{Aneumo Dataset and Geometric Preprocessing.}
We use the steady-flow subset of the publicly available Aneumo dataset.\cite{li2025aneumo,weidner2026inpainting} Its vascular geometries are derived from patient-specific intracranial aneurysm anatomies in the AneuX morphology database.\cite{juchler2022shape} The dataset authors generated synthetic anatomical variants through aneurysm-region removal, controlled local deformation, and geometric repair. Clinical and medical-imaging experts contributed to this process to support anatomical plausibility. The resulting geometries are patient-derived synthetic variants, rather than independently acquired patient anatomies or longitudinal observations of aneurysm progression.

\textbf{Mesh Generation and Reported Mesh Sensitivity.}
The Aneumo dataset authors generated unstructured polyhedral meshes using ANSYS Fluent Meshing 2023 R1, with \(10\) near-wall inflation layers and a growth ratio of \(1.2\).\cite{li2025aneumo} Their mesh-sensitivity assessment considered minimum cell sizes of \(0.25\), \(0.20\), \(0.15\), \(0.10\), and \(0.05~\mathrm{mm}\), corresponding to approximately \(150{,}000\), \(250{,}000\), \(330{,}000\), \(530{,}000\), and \(880{,}000\) cells, respectively. Velocity distributions and volume-averaged pressure were compared across these resolutions. The authors selected a minimum cell size of \(0.15~\mathrm{mm}\) as a balance between numerical accuracy and computational cost. These results document the mesh-sensitivity assessment of the source dataset; no independent mesh-refinement study was performed in the present work.

\textbf{CFD Setup and Vessel-Wall Assumptions.}
Blood was modeled as an incompressible Newtonian fluid with a dynamic viscosity of \(\mu=0.00345~\mathrm{Pa\cdot s}\) and a density of \(\rho=1050~\mathrm{kg/m^3}\). Three-dimensional simulations were performed using OpenFOAM v2312 with the transient incompressible solver \textit{icoFoam} and the PISO algorithm for pressure--velocity coupling. The vessel walls were treated as stationary and rigid, with a no-slip boundary condition. Consequently, the reference simulations describe flow within fixed vascular geometries and exclude wall compliance, deformation, and fluid--structure interaction. 

Eight constant inlet mass-flow rates were prescribed: \(0.0010\), \(0.0015\), \(0.0020\), \(0.0025\), \(0.0030\), \(0.0035\), \(0.00375\), and \(0.0040~\mathrm{kg/s}\). A gauge pressure of \(0~\mathrm{Pa}\) was imposed at the outlet. Time integration under these constant boundary conditions was used to approach approximately steady solutions, consistent with Eq.~\ref{eq:navier_stokes}. These fields support the steady pressure--velocity evaluation; the additional transient velocity experiments use the separate BenchAnXplore dataset.
%\textbf{CFD Setup and Vessel-Wall Assumptions.} The steady-flow reference data were obtained from the Aneumo dataset, in which blood was modeled as an incompressible Newtonian fluid with a dynamic viscosity of \(\mu=0.00345~\mathrm{Pa\cdot s}\) and a density of \(\rho=1050~\mathrm{kg/m^3}\). Three-dimensional simulations were performed using OpenFOAM v2312 with the transient incompressible solver \textit{icoFoam} and the PISO algorithm for pressure--velocity coupling. The vessel walls were assumed to be stationary and rigid, with a no-slip boundary condition, thereby excluding wall compliance, deformation, and fluid--structure interaction. Eight constant inlet mass-flow rates were prescribed: \(0.0010\), \(0.0015\), \(0.0020\), \(0.0025\), \(0.0030\), \(0.0035\), \(0.00375\), and \(0.0040~\mathrm{kg/s}\). A gauge pressure of \(0~\mathrm{Pa}\) was imposed at the outlet.

The Aneumo reference simulations used a fixed time step of \(\Delta t=1\times10^{-5}~\mathrm{s}\) over a duration of \(1~\mathrm{s}\), corresponding to \(100{,}000\) time steps, with a reported Courant number below unity. The transient solver \textit{icoFoam} advanced the flow under constant boundary conditions toward an approximately steady solution. The dataset authors reported that the velocity-component residuals decreased to approximately \(10^{-9}\) after \(20{,}000\) iterations, while the pressure residual remained below \(10^{-5}\). These observations document the reported solver convergence; they are not prescribed stopping tolerances and do not independently establish temporal stationarity or time-step independence. No separate time-step refinement study was reported. The resulting fields are used for the steady pressure--velocity benchmark, while cardiac-cycle dynamics are evaluated separately in the transient experiments.
%Although \textit{icoFoam} is a transient solver, time integration was used to approach an approximately steady solution under these time-independent boundary conditions, consistent with the steady-flow formulation in Eq.~\eqref{eq:navier_stokes}. The dataset authors used a fixed time step of \(\Delta t=1\times10^{-5}~\mathrm{s}\) over a simulation duration of \(1~\mathrm{s}\), corresponding to \(100{,}000\) time steps, and reported a Courant number below unity. They also reported that the velocity residuals decreased to approximately \(10^{-9}\) after \(20{,}000\) iterations, while the pressure residual remained below \(10^{-5}\). These residual levels indicate solver convergence but do not, by themselves, establish temporal stationarity, solution accuracy, or time-step independence. No separate time-step refinement study was reported. In the present study, these reference fields are used to evaluate approximately steady pressure--velocity prediction under prescribed inflow conditions; transient velocity-field prediction is evaluated separately using the BenchAnXplore dataset.

\textbf{Data preprocessing and training configuration.}
The dataset contains 500 vascular geometries, of which 400 are used for training and 100 for testing. Each geometry is paired with the eight prescribed inflow conditions, resulting in 3,200 training cases and 800 test cases, for a total of 4,000 CFD simulations. Each full vascular point cloud contains approximately \(2\times10^{5}\) points, including approximately \(10^{3}\) inlet points. Both the minimum and maximum values of pressure and velocity have been normalized to \((-1,1)\), based on the training dataset, and the test dataset has been processed in the same way. The spatial coordinates are scaled by a factor of 50 to improve numerical conditioning. During training, we randomly sample $\widetilde N_T=5000$, $\widetilde N_a=300$ points from the vascular domain and inlet boundary, respectively. The sampling sizes are selected based on the ablation study to balance prediction accuracy and computational cost.

The Point-ResKAN is implemented in PyTorch 2.5.0 and trained on an NVIDIA GeForce RTX 4090 GPU. The model is optimized using Adam with a weight decay of 0.05 and a batch size of 8 for 5,000 epochs. A predefined two-stage learning-rate schedule is used over \(5000\) training epochs. The learning rate is initially set to \(4\times10^{-4}\) and reduced to \(1\times10^{-4}\) at epoch \(2500\), allowing finer parameter updates during the second half of training. This transition is specified before training and is independent of the test-loss trajectory. The decrease in test loss and reduced fluctuations near epoch \(2500\) are consistent with the scheduled learning-rate reduction; this change in convergence behavior should therefore not be interpreted solely as an architectural advantage. We calculate the losses for the pressure and velocity components separately, multiply each loss (relative $L_2$ loss) by 100, and then add them together to obtain the total loss.

\textbf{Evaluation metrics.}
We evaluate prediction accuracy using the mean squared error (MSE), mean absolute error (MAE), and relative \(L_2\) error. For a test case containing \(N_{\mathrm{eval}}\) evaluation points, let $\boldsymbol y_n^{\mathrm{CFD}}=\left(P_n^{\mathrm{CFD}},\mathbf u_n^{\mathrm{CFD}}\right)$ and $\widehat{\boldsymbol y}_n=\left(\widehat P_n,\widehat{\mathbf u}_n\right)
\in\mathbb{R}^{d+1}$ denote the normalized CFD reference and Point-ResKAN prediction at the \(n\)-th point, respectively. The complete predicted field is obtained from
$\left(\widehat{\boldsymbol y}_n
\right)_{n=1}^{N_{\mathrm{eval}}}=
\mathcal{S}_{\theta}\left(\mathcal{X}_T,
\mathcal{Z}_{a,T}\right)$. The evaluation metrics are defined as
\begin{equation}
\begin{aligned}
\mathrm{MSE}
&=
\frac{1}{N_{\mathrm{eval}}}
\sum_{n=1}^{N_{\mathrm{eval}}}
\left\|
\boldsymbol y_n^{\mathrm{CFD}}
-
\widehat{\boldsymbol y}_n
\right\|_2^2,\\
\mathrm{MAE}
&=
\frac{1}{N_{\mathrm{eval}}}
\sum_{n=1}^{N_{\mathrm{eval}}}
\left\|
\boldsymbol y_n^{\mathrm{CFD}}
-
\widehat{\boldsymbol y}_n
\right\|_1,\\
\mathrm{Relative}\ L_2
&=
\left(
\frac{
\displaystyle
\sum_{n=1}^{N_{\mathrm{eval}}}
\left\|
\boldsymbol y_n^{\mathrm{CFD}}
-
\widehat{\boldsymbol y}_n
\right\|_2^2
}{
\displaystyle
\sum_{n=1}^{N_{\mathrm{eval}}}
\left\|
\boldsymbol y_n^{\mathrm{CFD}}
\right\|_2^2
}
\right)^{1/2},\\
\mathrm{RMSE}
&=
\left(
\frac{1}{N_{\mathrm{eval}}}
\sum_{n=1}^{N_{\mathrm{eval}}}
\left\|
\mathbf{y}_n^{\mathrm{CFD}}
-
\widehat{\mathbf{y}}_n
\right\|_2^2
\right)^{1/2}.
\end{aligned}
\label{eq:evaluation_metrics}
\end{equation}
Because pressure and velocity have different physical units, the
aggregate metrics are computed using their normalized channels.

%%%%%%%%%%%%
%%%%%%%%%%%%%%%%%%%%%%%%%%%%%%%模型配置
\subsection{\label{sec:level-3.2}Comparison methods}

We compare the proposed Point-ResKAN with representative neural-operator and point-cloud-based baselines, including KA-PointNet,\cite{kashefi2025kolmogorov} PointNet,\cite{qi2017pointnet} Fourier Neural Operator (FNO),\cite{li2020fourier} and DeepONet.\cite{lu2021learning} The comparison is conducted on test cases with vascular geometries that are not included in the training set, thereby evaluating the ability of each method to generalize across patient-specific vascular geometries.

\begin{figure*}
 \centering
 \includegraphics[width=0.9\linewidth]{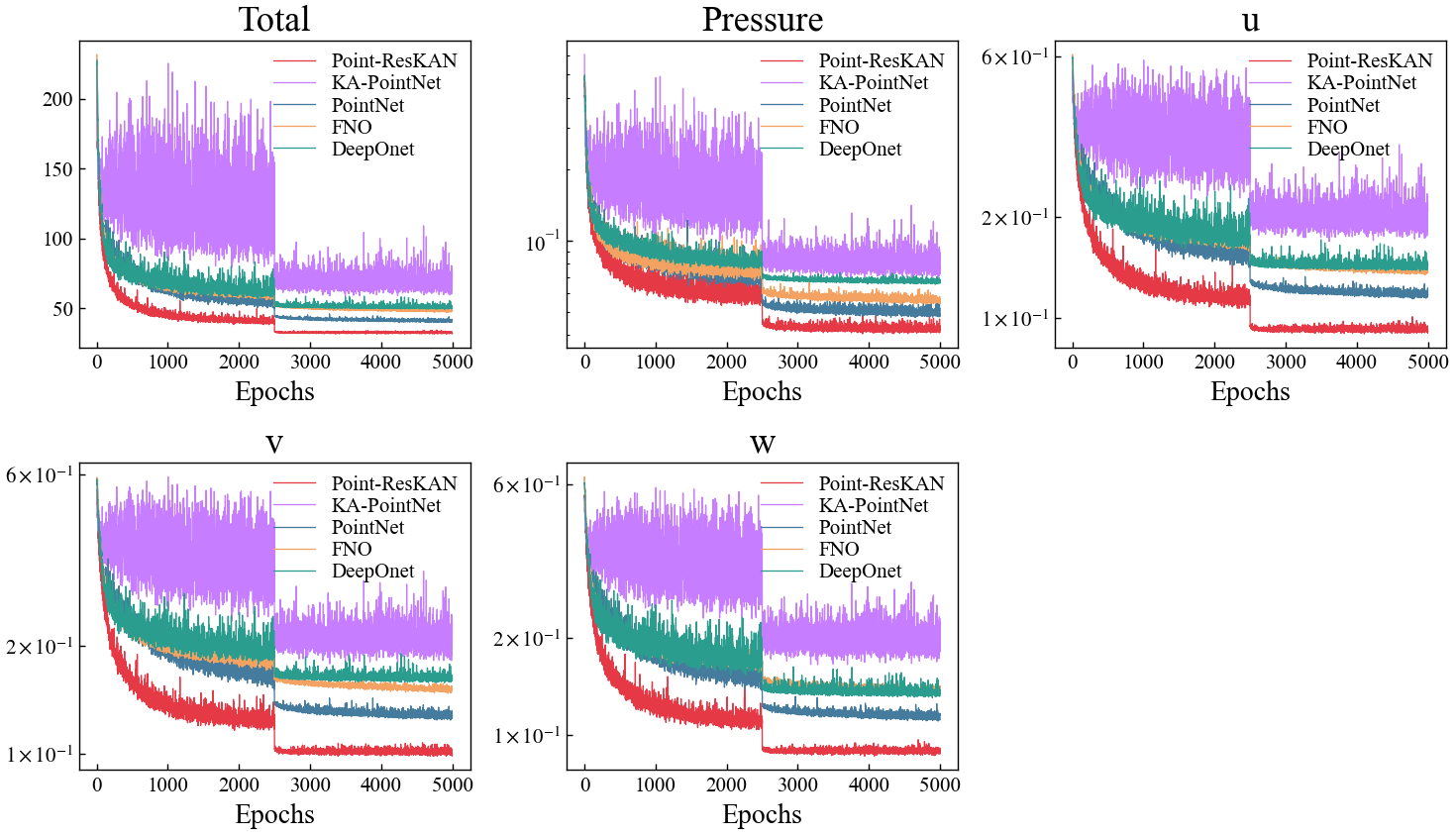}
 \caption{Comparison of test-loss trajectories across different surrogate models.}
 \label{fig:3}
\end{figure*}

\textbf{Baseline configurations.}
For a fair comparison, all baseline models use the same training and test splits, input normalization, point-sampling strategy, batch size, number of training epochs, and learning-rate schedule as Point-ResKAN, unless otherwise stated. Their architecture-specific configurations are as follows:
\begin{itemize}
    \item \textbf{DeepONet:}
    The trunk network encodes the spatial query points, whereas the branch network encodes the sampled inflow condition. Both networks consist of eight fully connected layers, with 256 hidden units in each layer.
    \item \textbf{FNO:}
    Because the hemodynamic data are represented on unstructured point clouds, we use a point-cloud-adapted FNO baseline. In the lifting layer, max-pooling is applied to the input boundary features, which are then concatenated with the pointwise local features. The latent representation is then processed by four Fourier layers with 32 retained modes and a channel width of 128.
    \item \textbf{PointNet:}
    The layer widths of the pointwise and global geometry mappings are
    $[3,64,128,128,256,256]$ and  $[256,256,256,512]$,
    respectively. The inflow encoder uses the layer widths
    $[7,128,128,256,256]$.
    \item \textbf{KA-PointNet:}
    The raw input features are first mapped to a 32-dimensional latent representation by a fully connected (FC) layer. The subsequent Jacobi-based KAN mappings use layer widths $[32,64,128,256]$ and $[256,512]$ for pointwise and global geometric encoding, respectively, while the inflow branch uses $[32,256]$. The edge functions are parameterized by Jacobi polynomials of degree \(n=2\), with \(\alpha=\beta=1.0\).
    \item \textbf{Point-ResKAN:}
    Point-ResKAN uses the same latent dimensions, Jacobi-polynomial degree, and basis parameters as KA-PointNet. The principal architectural difference is the learnable affine residual branch introduced in each ResKAN layer. This matched configuration allows the performance gain associated with the residual construction to be evaluated independently of the KAN width and polynomial basis. Among these, all decoders uses the FC layer widths $[1024,512,256,128,4]$.
\end{itemize}

\subsection{\label{sec:level-3.3}Results on unseen vascular geometries}
    
\textbf{Convergence behavior and training stability.}
Fig.~\ref{fig:3} compares the evolution of the test loss during training. Point-ResKAN decreases the loss more rapidly and exhibits smaller oscillations than the baseline models, whereas KA-PointNet converges more slowly and reaches a higher final test error. The decrease in test loss and its fluctuations around epoch 2,500 is consistent with the transition to the lower learning rate used for refinement. These results are consistent with the role of the affine residual branch in providing an additional path for feature and gradient propagation. By combining this direct affine path with the nonlinear Jacobi-based KAN mapping, ResKAN facilitates the optimization of stacked KAN layers and improves training stability on large-scale unstructured point clouds. This design enables Point-ResKAN to learn the nonlinear dependence of the pressure--velocity fields on vascular geometry and inflow conditions more effectively.

\begin{table}[htbp]
\centering
\caption{Performance comparison among different algorithms.}
\begin{tabular}{lccccc}
\hline\hline
\text{Algorithm} & \text{Parameter} & \text{MSE} & \text{MAE} & \text{$L_2$} \\
\midrule
\multirow{4}{*}{DeepONet} & Pressure & 0.002675 & 0.021517 & 0.069806 \\
                           & u        & 0.015127 & 0.061780 & 0.136695 \\
                           & v        & 0.014505 & 0.062002 & 0.156103 \\
                           & w        & 0.015963 & 0.065941 & 0.130533 \\
\midrule
\multirow{4}{*}{FNO} & Pressure & 0.003474 & 0.020959 & 0.063389 \\
                           & u        & 0.015182 & 0.063781 & 0.139776 \\
                           & v        & 0.01449 & 0.062125 & 0.157581 \\
                           & w        & 0.020622 & 0.072465 & 0.138586 \\
\midrule
\multirow{4}{*}{PointNet} & Pressure & 0.002299 & 0.019574 & 0.059841 \\
                           & u        & 0.013850 & 0.063454 & 0.132533 \\
                           & v        & 0.010429 & 0.058064 & 0.140461 \\
                           & w        & 0.014228 & 0.066309 & 0.125355 \\
\midrule
\multirow{4}{*}{KA-PointNet} & Pressure & 0.003570 & 0.027461 & 0.086670 \\
                           & u        & 0.027981 & 0.092445 & 0.198724 \\
                           & v        & 0.025814 & 0.087354 & 0.207621 \\
                           & w        & 0.032417 & 0.097627 & 0.188846 \\
\midrule
\multirow{4}{*}{Point-ResKAN}  & Pressure & 0.001097 & 0.014036 & 0.045980 \\
                           & u        & 0.007405 & 0.042048 & 0.096473 \\
                           & v        & 0.006132 & 0.038892 & 0.105846 \\
                           & w        & 0.008474 & 0.044788 & 0.090319 \\
\hline\hline
\end{tabular}
\label{table:2}
\end{table}

\begin{table}[htbp]
\centering
\footnotesize
\caption{ Dimensional prediction errors for Point-ResKAN over the complete test set, averaged across cases with equal weighting. Pressure MSE is in \(\mathrm{Pa^2}\); pressure MAE and maximum absolute error are in \(\mathrm{Pa}\). Velocity-component MSE is in \(\mathrm{m^2/s^2}\); velocity-component MAE and maximum absolute error are in \(\mathrm{m/s}\). The last column is the mean of the case-wise maximum absolute errors.}
\label{table:mae}
\begin{tabular}{lccc}
\hline\hline
Quantity & Mean case-wise & Mean case-wise & Mean case-wise \\
         & MSE            & MAE            & maximum absolute error \\
\hline
Pressure & 0.000171 & 0.005334 & 0.058382 \\
\(u\)    & 0.000863 & 0.014470 & 0.066117 \\
\(v\)    & 0.000746 & 0.013476 & 0.058737 \\
\(w\)    & 0.000964 & 0.015190 & 0.065882 \\
\hline\hline
\end{tabular}
\end{table}

\textbf{Predictive accuracy and benchmark comparison.} As summarized in Table~\ref{table:2} and Table~\ref{table:mae}, Point-ResKAN achieves the lowest MSE, MAE, and relative \(L_2\) error for all predicted quantities, including pressure \(P\) and the velocity components \((u,v,w)\). For pressure, Point-ResKAN attains a relative \(L_2\) error of \(0.04598\), representing reductions of \(23.2\%\), \(46.9\%\), \(27.5\%\), and \(34.1\%\) relative to PointNet, KA-PointNet, FNO, and DeepONet, respectively. The corresponding relative \(L_2\) errors for \(u\), \(v\), and \(w\) are \(0.0965\), \(0.1058\), and \(0.0903\), respectively, each lower than that of the competing methods. These consistent improvements across pressure and velocity indicate that the affine residual paths enhance the Jacobi-based KAN mappings without sacrificing their nonlinear approximation capability. Overall, Point-ResKAN provides more accurate pressure--velocity predictions for vascular geometries not encountered during training.

\textbf{Dimensional Prediction Errors.} Table~\ref{table:mae} supplements the normalized comparisons with dimensional errors for Point-ResKAN over the complete test set. After inverse normalization, MSE, MAE, and maximum absolute error are computed separately for pressure and each velocity component over all available points in each case. These case-wise statistics are then averaged with equal case weighting. Pressure MAE and maximum absolute error are expressed in \(\mathrm{Pa}\), and the corresponding velocity-component errors in \(\mathrm{m/s}\). MSE has units of \(\mathrm{Pa^2}\) and \(\mathrm{m^2/s^2}\), respectively. The maximum-error column reports the mean of the case-wise maxima, not the global maximum across the test set. These point-based statistics are complemented by spatial absolute-error maps; they do not represent volume-weighted spatial integrals.

\begin{figure*}
 \centering
 \includegraphics[width=0.85\linewidth]{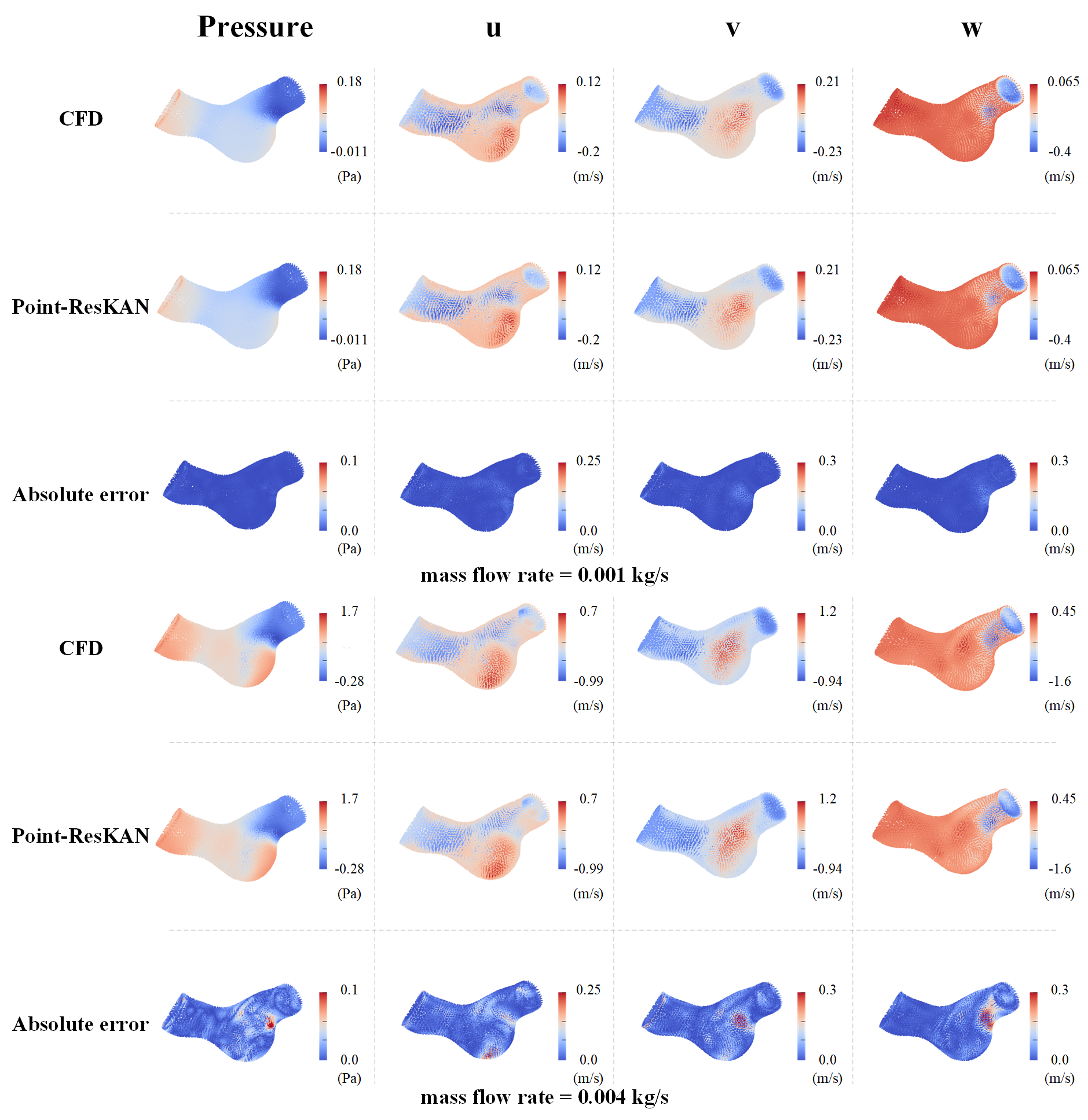}
 \caption{Comparison of flow field distribution by CFD and Point-ResKAN. The error shows pointwise absolute error between the Point-ResKAN and CFD results after inverse normalization. Pressure errors are expressed in Pa, and velocity-component errors are expressed in m/s.}
 \label{fig:5}
\end{figure*}

\begin{figure*}
 \centering
 \includegraphics[width=0.85\linewidth]{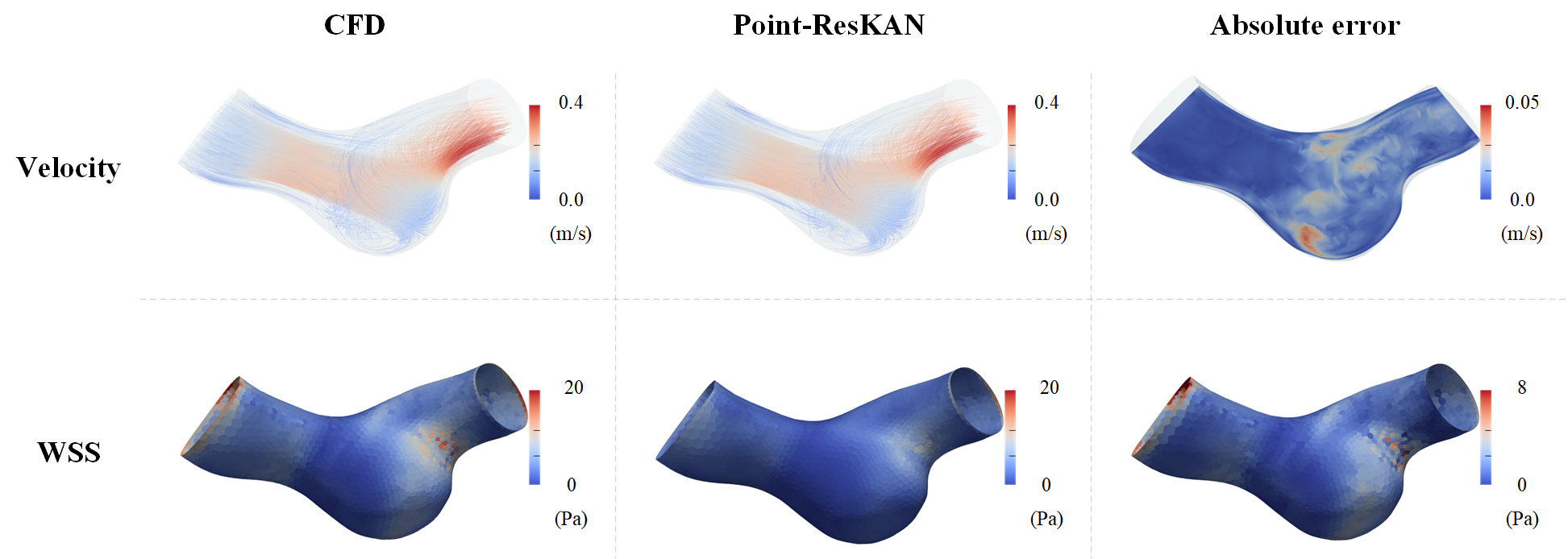}
 \caption{Comparison of streamlines and reconstructed wall shear stress (WSS) magnitude from the CFD reference and Point-ResKAN velocity fields for a steady-flow test case at an inlet mass-flow rate of \(0.001~\mathrm{kg/s}\). Absolute-error maps are shown for velocity magnitude and WSS magnitude. Corresponding reference and predicted quantities share identical color scales. Velocity and WSS are expressed in \(\mathrm{m/s}\) and \(\mathrm{Pa}\), respectively.}
 \label{fig:x1}
\end{figure*}

\begin{figure*}
 \centering
 \includegraphics[width=0.85\linewidth]{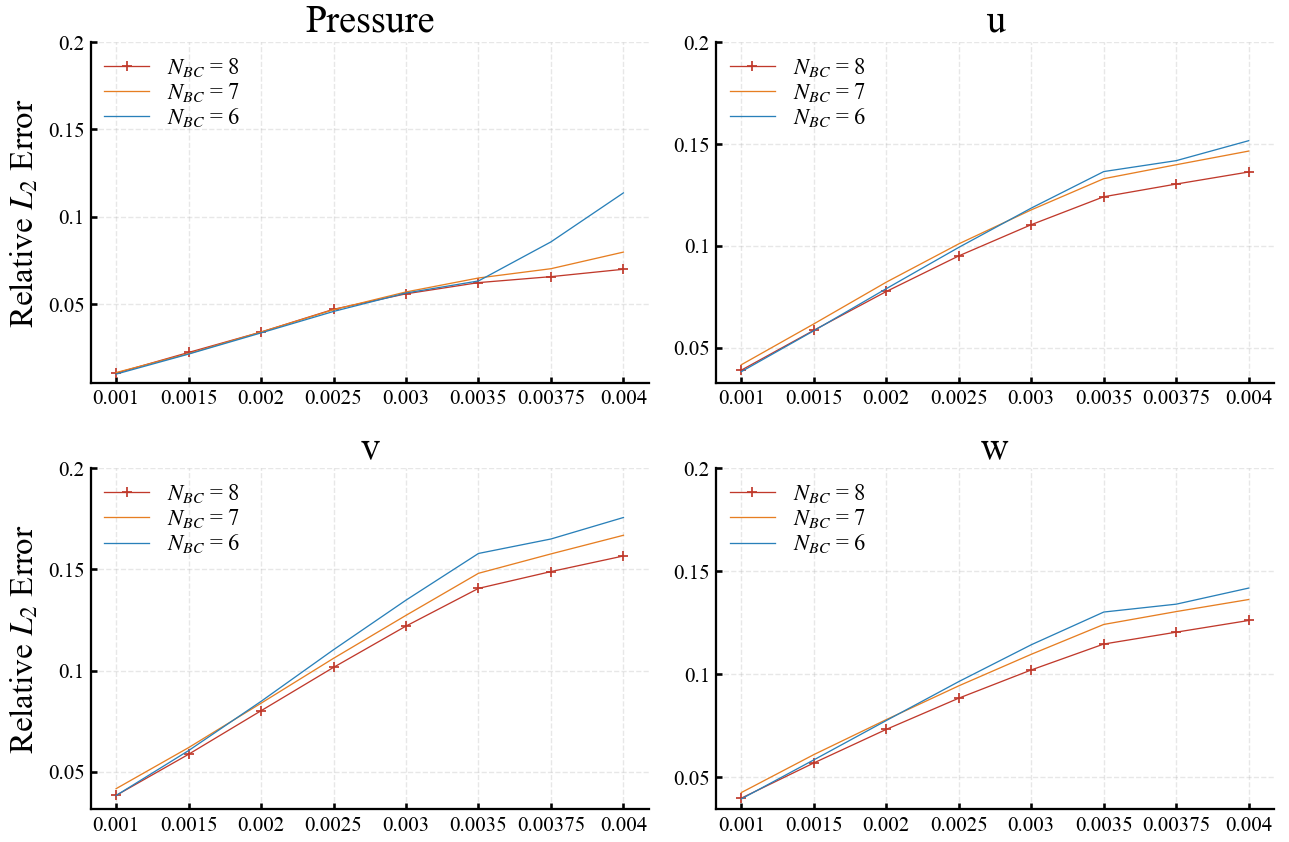}
 \caption{Comparison of the relative $L_2$ errors for Point-ResKAN under unseen inflow boundary conditions. The OOD  are \(0.004~\mathrm{kg/s}\) for \(N_{\mathrm{bc}}=7\), and \(0.00375\) and \(0.004~\mathrm{kg/s}\) for \(N_{\mathrm{bc}}=6\).}
 \label{fig:6}
\end{figure*}

\begin{figure*}
 \centering
 \includegraphics[width=0.85\linewidth]{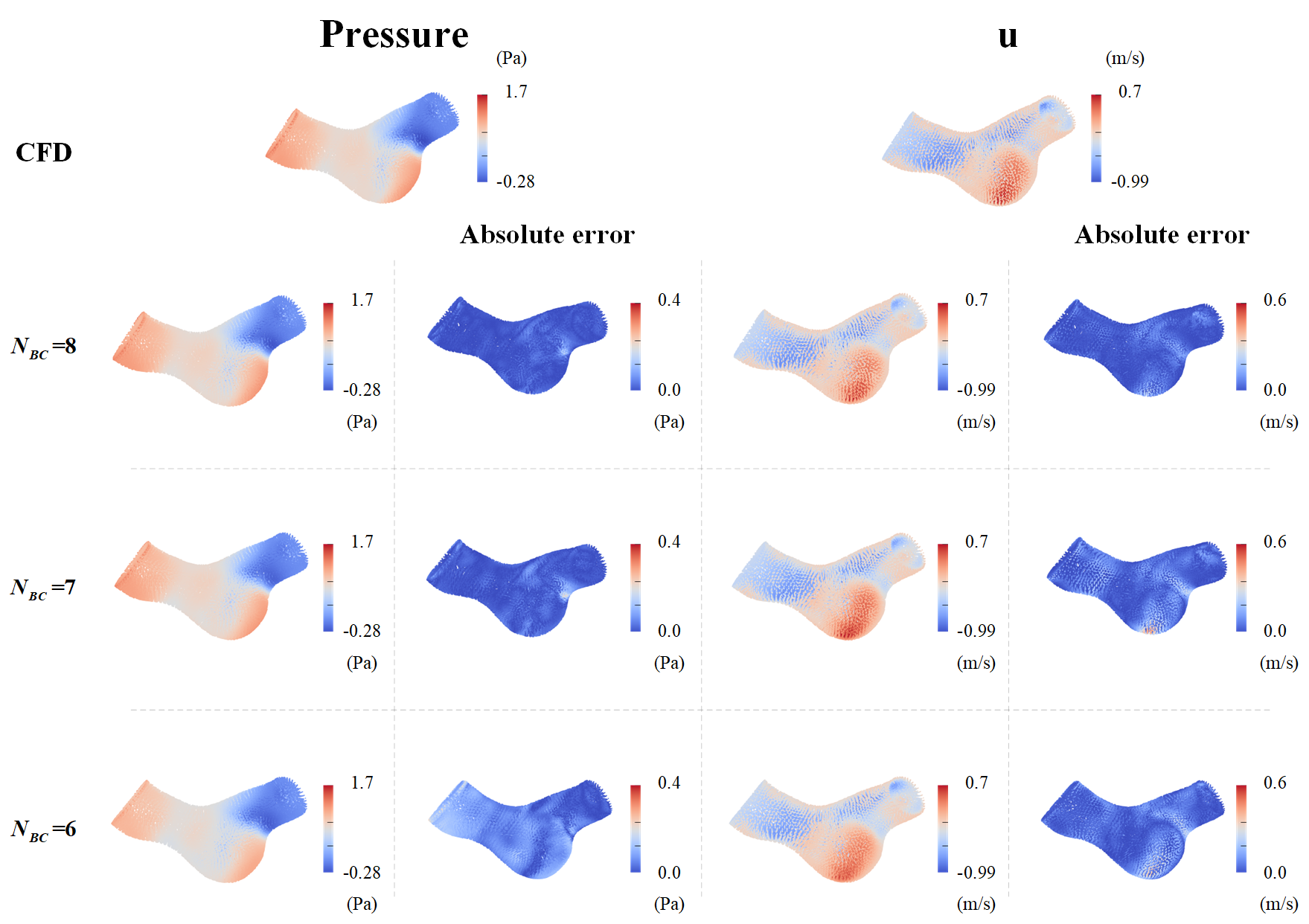}
 \caption{Comparison of CFD and Point-ResKAN flow-field distributions under variable inflow boundary conditions.}
 \label{fig:7}
\end{figure*}

\textbf{Point Sampling and Training-Sample Sensitivity.} Training uses randomly selected subsets of vascular-domain and inlet points, renewed throughout training. Testing uses the complete available domain and inlet point clouds. Pressure and velocity are predicted directly at the vascular coordinates and compared with the corresponding CFD results, without sparse-to-dense interpolation. We assess the training sample count using \(2000\), \(5000\), and \(15000\) domain points, with \(300\) inlet points and otherwise unchanged training settings. All configurations are evaluated on the same complete test point clouds. Table~\ref{table:point} shows that increasing the domain count from \(2000\) to \(5000\) improves all reported field errors. Increasing it to \(15000\) yields mixed relative changes below \(3.5\%\), while increasing training time from \(13.5\) to \(37.4~\mathrm{s}\) per epoch. We therefore retain \(5000\) domain points as an accuracy--cost compromise. Random sampling does not explicitly prioritize near-wall or high-gradient regions. The observed global-error sensitivity should therefore be distinguished from local spatial convergence, which would require additional evaluation of sampling distributions and region-specific errors.

\textbf{Model Capacity and Training Cost.} Table~\ref{table:cost} reports trainable-parameter counts, training time per epoch, saved model-file size, and observed GPU memory usage during training for the Aneumo experiments. Measurements were obtained on the same NVIDIA GeForce RTX 4090 GPU with a batch size of \(16\), using the same training dataset and number of sampled points per epoch. Hyperparameters were tuned separately for each model based on validation performance using a common data partition. Point-ResKAN has \(1.42\) million trainable parameters and requires \(13.5~\mathrm{s}\) per epoch, compared with \(1.39\) million parameters and \(11.5~\mathrm{s}\) for KA-PointNet. The corresponding GPU memory usage is approximately \(4.0\) and \(3.9~\mathrm{GB}\), respectively. Together with the prediction errors, these measurements characterize the accuracy--cost trade-off and make differences in model capacity explicit. Saved model-file size is reported as a storage metric rather than a measure of computational complexity.

\begin{table}[htbp]
\centering
\caption{Comparison of accuracy and training time per epoch(s) among different sampling schemes.}
\begin{tabular}{lccccc}
\hline\hline
\text{Sampling} & Pressure & u & v & w& Training(s)\\
\hline
2000 & 0.046214 & 0.108419 & 0.117979 & 0.104047 &9.7\\
5000 & 0.045980 & 0.096473 & 0.105846 & 0.090319 &13.5\\
15000& 0.044378 & 0.097738 & 0.104918 & 0.090688 &37.4\\
\hline\hline
\end{tabular}
\label{table:point}
\end{table}

\begin{table}[htbp]
\centering
\footnotesize
\caption{Model capacity and training cost for the Aneumo experiments. GPU memory denotes observed usage during training; parameter counts are expressed in millions.}
\label{table:cost}
\begin{tabular}{lcccc}
\hline\hline
Model & Time/epoch & Model file & GPU memory & Parameters \\
      & (s)        & (MB)       & (GB)       & (M) \\
\hline
DeepONet     & 11.4 & 4.7  & 2.4 & 1.24 \\
FNO          & 9.4  & 20.9 & 2.1 & 5.49 \\
PointNet     & 10.5 & 6.8  & 3.2 & 1.78 \\
KA-PointNet  & 11.5 & 5.3  & 3.9 & 1.39 \\
Point-ResKAN & 13.5 & 5.5  & 4.0 & 1.42 \\
\hline\hline\\
\end{tabular}
\end{table}

\begin{table}[htbp]
\centering
%\small  % 或 
%\fontsize{10pt}{12pt}\selectfont
\footnotesize
\setlength{\tabcolsep}{2pt}  % 减小列间距，默认约为6pt
\caption{Performance comparison of different algorithms on the pressure field of the test dataset.}
%\begin{tabular*}{\linewidth}{@{\extracolsep{\fill}} l *{5}{c} @{}}
\begin{tabular}{l*{5}{c}}
\hline\hline
\text{Case} & \text{DeepONet} & \text{ FNO} & \text{PointNet} & \text{KA-PointNet} & \text{Point-ResKAN} \\
\midrule
1  & 0.178285 & 0.134681 & 0.111039 & 0.140581 & 0.107675 \\
2  & 0.104431 & 0.073446 & 0.069407 & 0.084298 & 0.070101 \\
3 & 0.091308 & 0.058438 & 0.046549 & 0.147351 & 0.031966 \\
4 & 0.071083 & 0.079107 & 0.056475 & 0.085251 & 0.059112 \\
5 & 0.025805 & 0.037072 & 0.025481 & 0.049545 & 0.029349 \\
6 & 0.015191 & 0.020287 & 0.013891 & 0.026013 & 0.011144 \\
7 & 0.053283 & 0.043411 & 0.045619 & 0.113049 & 0.034415 \\
8 & 0.038757 & 0.044028 & 0.039483 & 0.108211 & 0.034862 \\
9 & 0.127913 & 0.216424 & 0.116389 & 0.165567 & 0.104073 \\
10 & 0.017751 & 0.016483 & 0.016878 & 0.040763 & 0.010510 \\
11 & 0.022428 & 0.023901 & 0.024322 & 0.029196 & 0.015181 \\
12 & 0.039548 & 0.032439 & 0.029987 & 0.039591 & 0.021323 \\
13 & 0.047248 & 0.054165 & 0.043014 & 0.083450 & 0.043325 \\
14 & 0.019081 & 0.020935 & 0.017652 & 0.032057 & 0.015601 \\
15 & 0.013435 & 0.014368 & 0.016451 & 0.035121 & 0.013074 \\
16 & 0.085398 & 0.068728 & 0.057989 & 0.078583 & 0.048814 \\
17 & 0.163016 & 0.130184 & 0.149642 & 0.215915 & 0.120946 \\
18 & 0.094997 & 0.052078 & 0.046509 & 0.077771 & 0.028669 \\
19 & 0.128245 & 0.103369 & 0.090215 & 0.113464 & 0.084691 \\
20 & 0.058917 & 0.044232 & 0.045923 & 0.067624 & 0.034764 \\
\midrule
Mean & 0.069806 & 0.063389 & 0.053146 & 0.086670 & \textbf{0.045980} \\
Std & $\pm 0.04909$ & $\pm 0.04888$ & $\pm 0.03634$ & $\pm 0.05012$ & $\pm \textbf{0.03336}$ \\
\hline\hline
\end{tabular}
\label{table:3}
\end{table}

\textbf{Robustness to geometric heterogeneity.}
We randomly selected 20 cases from the test datasets to display the accuracy of the algorithm. As reported in Table~\ref{table:3}, Point-ResKAN achieves the smallest standard deviation of the case-wise prediction errors among the compared methods, with a value of \(0.03336\). This result indicates more consistent predictive performance across the test geometries. However, all methods exhibit increased errors for geometrically complex aneurysm cases, particularly cases 1 and 17. These observations suggest that the affine residual branches improve the stability and representation capability of the Jacobi-based KAN mappings, while pronounced morphological heterogeneity remains a major challenge for point-cloud-based surrogate modelling.

\begin{figure*}
 \centering
 \includegraphics[width=0.85\linewidth]{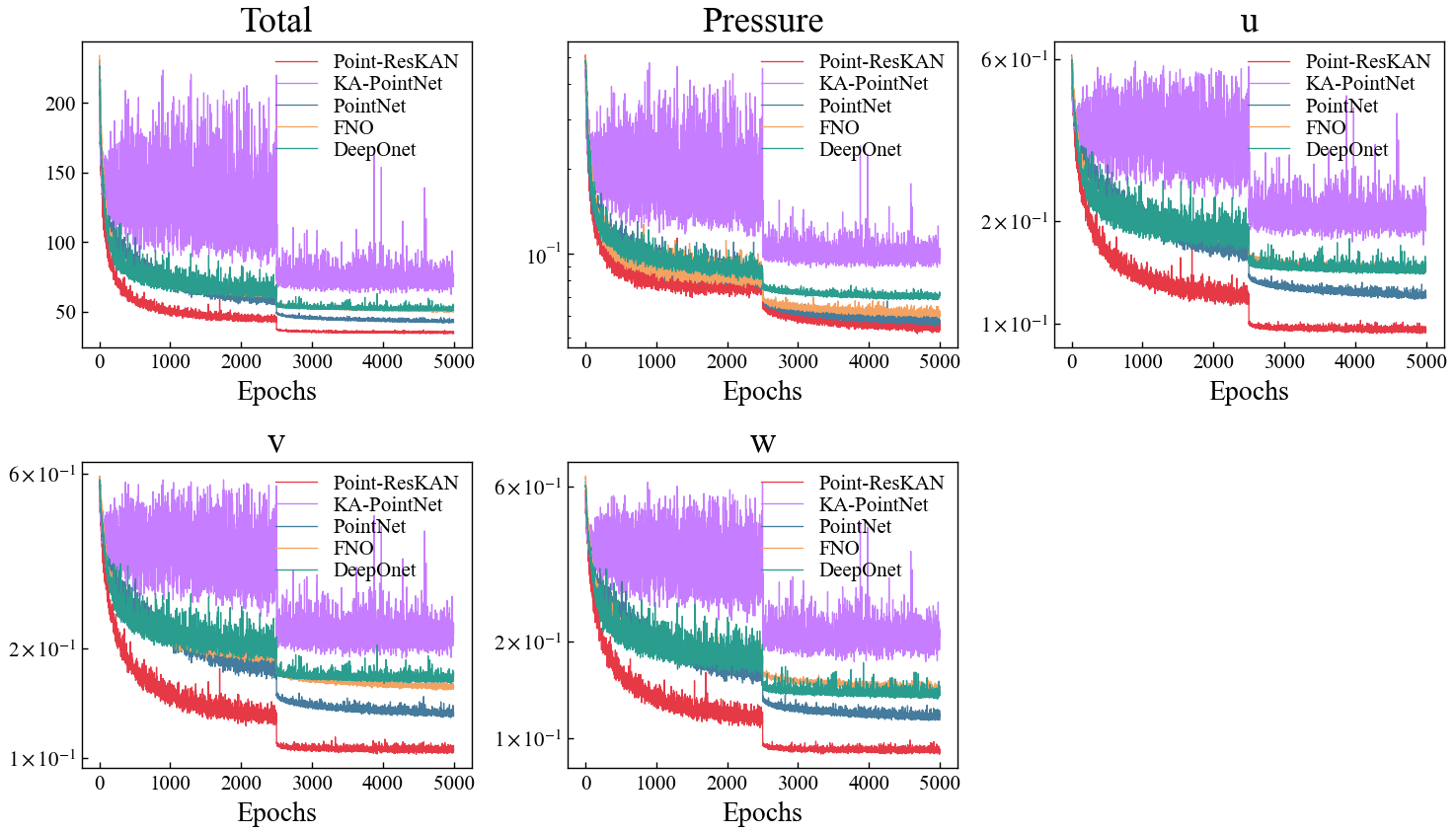}
 \caption{Comparative analysis of testing loss trajectories under specific inflow boundary conditions.}
 \label{fig:8}
\end{figure*}
\begin{figure*}
 \centering
 \includegraphics[width=0.85\linewidth]{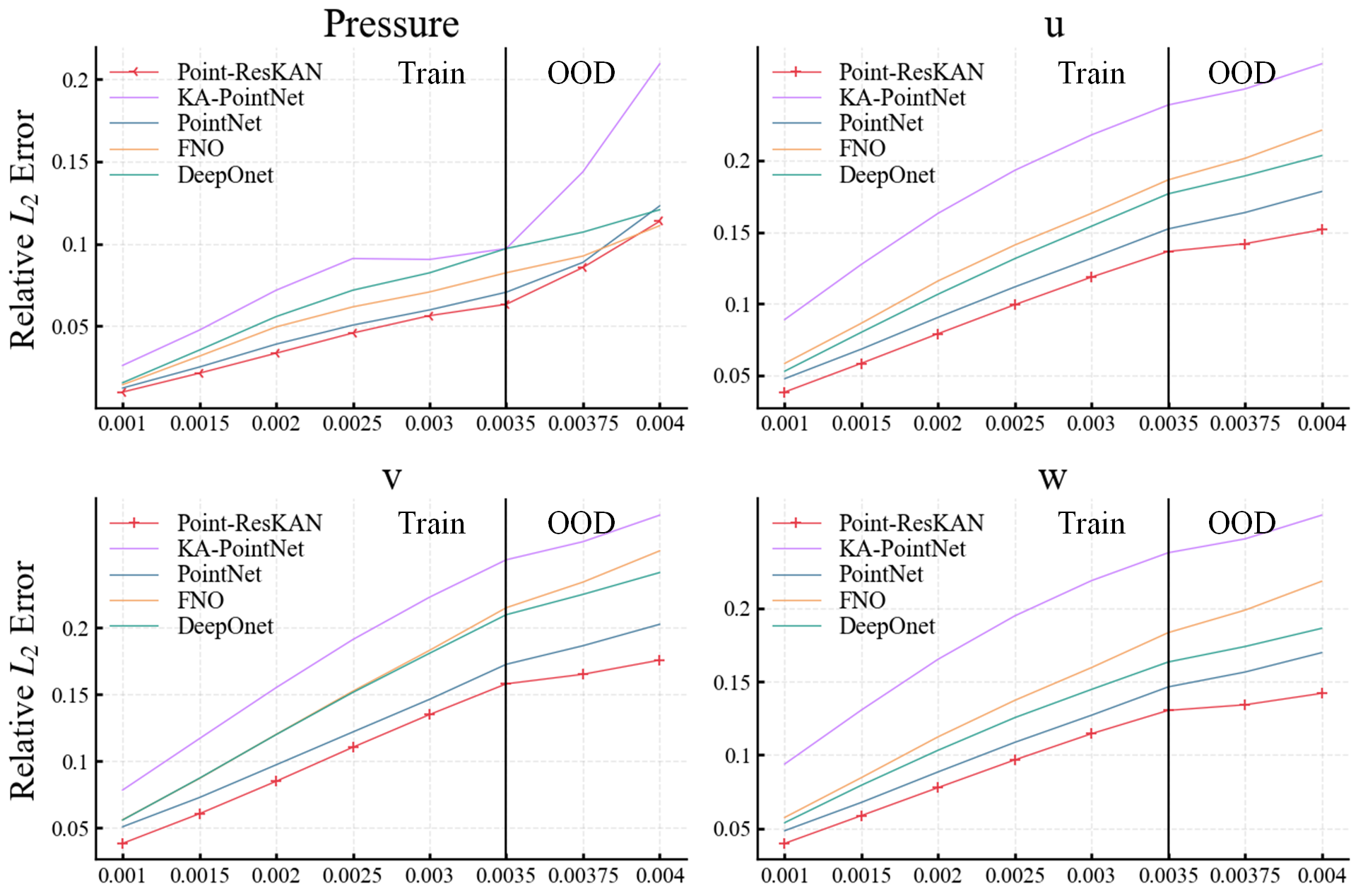}
 \caption{Comparison of the relative $L_2$ error for different algorithms under specific inflow boundary conditions.}
 \label{fig:9}
\end{figure*}
\begin{figure*}
 \centering
 \includegraphics[width=0.9\linewidth]{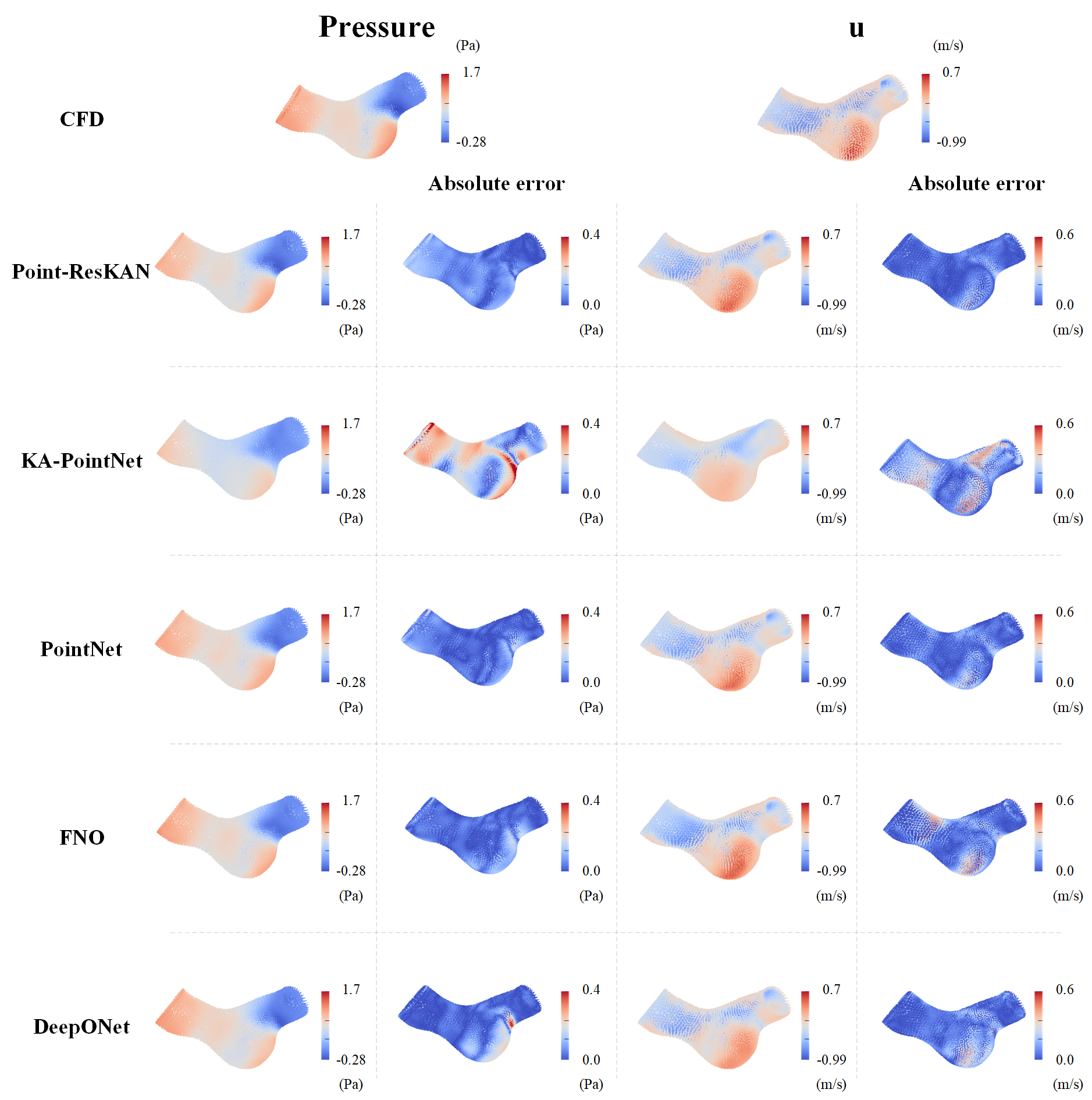}
 \caption{Comparison of flow field distribution for the different surrogate models under specific inflow boundary conditions.}
 \label{fig:10}
\end{figure*}

\textbf{Computational efficiency and potential application.}
Point-ResKAN substantially reduces the computational cost of hemodynamic prediction relative to the reference CFD simulations. For the intracranial aneurysm cases considered in this study, each CFD simulation requires approximately \(45\) minutes to \(2\) hours on \(24\) CPU cores. Once trained, Point-ResKAN predicts the pressure--velocity values at all sampled points in less than \(1\) second, excluding the one-time training cost. This corresponds to an inference speedup of more than three orders of magnitude and enables rapid evaluation of multiple inflow conditions for a given vascular geometry. The speedup is defined as the ratio of CFD simulation time per case to the inference time per case after training. Relative to the evaluated baselines, Point-ResKAN reduces the prediction errors while retaining rapid inference, with its training time, model size, and GPU memory requirements to quantify the associated accuracy–cost tradeoff.

\textbf{Qualitative comparison and physical interpretation.}
Fig.~\ref{fig:5} compares the CFD reference fields with the Point-ResKAN predictions for inlet mass-flow rates of \(0.001~\mathrm{kg/s}\) and \(0.004~\mathrm{kg/s}\). The CFD solutions exhibit elevated pressure near the inflow-jet impingement region, together with recirculating flow and low-velocity zones within the aneurysm sac. Such flow features are relevant to aneurysm hemodynamics, in which jet impingement, wall shear stress, oscillatory shear, and local wall properties may interact in determining rupture risk.\cite{cebral2015analysis}

\textbf{Flow Patterns and Reconstructed Wall Shear Stress.} Fig.~\ref{fig:x1} compares streamlines and reconstructed WSS magnitudes obtained from the CFD reference and Point-ResKAN velocity fields for a steady-flow test case at an inlet mass-flow rate of \(0.001~\mathrm{kg/s}\). Absolute-error maps are provided for velocity magnitude and WSS magnitude, and corresponding reference and predicted quantities use identical color scales. Point-ResKAN reproduces the overall streamline patterns in this case, while the reconstructed WSS maps reveal localized discrepancies near the vessel junction and end regions. These comparisons extend the evaluation beyond global field errors by examining spatial flow organization and a derived near-wall hemodynamic quantity.

Point-ResKAN reproduces the principal spatial patterns of the CFD solutions, including the impinging inflow jet near the aneurysm neck and the dominant recirculation structures within the aneurysm lumen. The largest discrepancies occur mainly near regions of high curvature, steep velocity gradients, and concentrated vortical flow, and become more pronounced at the higher mass-flow rate of \(0.004~\mathrm{kg/s}\). Overall, the qualitative results demonstrate that Point-ResKAN preserves the main pressure and velocity characteristics of the reference solutions while providing rapid hemodynamic predictions conditioned on the prescribed inflow.
\vspace{-1em}

\subsection{Results on unseen inflow conditions}
\label{sec:inflow_generalization}

We further evaluate prediction at selected steady inlet mass-flow rates excluded from training. The complete dataset contains eight prescribed inlet mass-flow rates. We vary the number of inflow conditions used for training as $N_{\mathrm{BC}}\in\{8,7,6\}$. The training and out-of-distribution (OOD) ranges for unseen inflow conditions are shown in Table~\ref{table:ood}. The vascular geometries used for testing remain disjoint from those in the training set. This experiment therefore evaluates whether the learned point-cloud surrogate can assess limited extrapolation above the training inflow range under the same steady-flow assumptions.

\textbf{Sensitivity to Training Inflow Coverage.} Fig.~\ref{fig:6} compares prediction errors for models trained with different numbers of discrete inflow conditions, \(N_{\mathrm{BC}}\). Table~\ref{table:ood} identifies the withheld rates above each training range. Errors generally increase as training inflow coverage is reduced, particularly for pressure at \(0.004~\mathrm{kg/s}\) when \(N_{\mathrm{BC}}=6\). The spatial error maps in Figure 8 complement these aggregate metrics by showing where prediction discrepancies occur. These experiments quantify performance at selected extrapolation conditions and show the importance of training inflow coverage. Because reducing \(N_{\mathrm{BC}}\) changes both the number of training conditions and the upper training bound, the results reflect their combined effects. They are interpreted as discrete extrapolation tests, rather than evidence of continuous interpolation accuracy or unrestricted generalization across inflow conditions.

\begin{table}[htbp]
\centering
\footnotesize
\caption{Training mass-flow-rate bounds and withheld extrapolation conditions. The bounds summarize discrete training rates and do not imply continuous sampling.}
\label{table:ood}
\begin{tabular}{lcc}
\hline\hline
Configuration & Training bounds (\(\mathrm{kg/s}\))
              & Extrapolation rates (\(\mathrm{kg/s}\)) \\
\hline
\(N_{\mathrm{BC}}=8\) & \([0.001,\,0.004]\)   & --- \\
\(N_{\mathrm{BC}}=7\) & \([0.001,\,0.00375]\) & \(0.004\) \\
\(N_{\mathrm{BC}}=6\) & \([0.001,\,0.0035]\)  & \(0.00375,\;0.004\) \\
\hline\hline
\end{tabular}
\end{table}

\textbf{Generalization relative to baseline methods.}
To further evaluate generalization to unseen inflow conditions, selected inlet mass-flow rates, including \(0.00375~\mathrm{kg/s}\) and \(0.004~\mathrm{kg/s}\), are excluded from training and used only for testing. The comparison includes DeepONet, FNO, PointNet, KA-PointNet, and Point-ResKAN. As shown in Fig.~\ref{fig:8}, Point-ResKAN reduces the test loss more rapidly than the baseline models and exhibits smaller oscillations during the later stages of training. In particular, its smoother trajectory relative to KA-PointNet is consistent with the affine residual branch providing an additional feature- and gradient-propagation path for the Jacobi-based KAN layers. As reported in Table~\ref{table:4}, Point-ResKAN achieves a relative \(L_2\) error of \(0.0562\) for pressure, corresponding to reductions of \(23.3\%\), \(12.5\%\), \(8.9\%\), and \(42.2\%\) relative to DeepONet, FNO, PointNet, and KA-PointNet, respectively. Point-ResKAN also obtains the lowest relative \(L_2\) errors for all velocity components. These results demonstrate that the proposed residual KAN architecture more accurately predicts pressure--velocity fields at inflow conditions not represented in the training set.

\begin{table}[htbp]
\centering
\caption{Comparison of the mean relative $L_2$ error for different algorithms under specific inflow boundary conditions.}
%\begin{tabular}{l S[table-format=0.6] S[table-format=0.6] S[table-format=0.6] S[table-format=0.6]}
\begin{tabular}{lcccc}
\hline\hline
\text{Model} & \text{Pressure} & \text{u} & \text{v} & \text{w} \\
\hline
DeepONet & 0.073262 & 0.136797 & 0.158891 & 0.128684 \\
FNO      & 0.064278 & 0.146699 & 0.163147 & 0.143864 \\
PointNet & 0.061689 & 0.128008 & 0.121262 & 0.119014 \\
KA-PointNet & 0.097189 & 0.193287 & 0.195528 & 0.193922 \\
\hline
Point-ResKAN & \textbf{0.056212} & \textbf{0.102969} & \textbf{0.115899} & \textbf{0.099136} \\
\hline\hline
\end{tabular}
\label{table:4}
\end{table}

\textbf{Generalization to higher-flow conditions.}
Fig.~\ref{fig:9} shows the variation in prediction error as the inlet mass-flow rate increases from \(0.001\) to \(0.004~\mathrm{kg/s}\). For all methods, the relative error generally increases at higher flow rates, where stronger convective effects and steeper spatial gradients produce a more nonlinear dependence of the pressure--velocity field on the inflow condition. Point-ResKAN exhibits a smaller error increase than the baseline methods, indicating greater robustness to changes in the inflow regime. These results suggest that the affine residual paths enhance the capacity of the Jacobi-based KAN mappings to represent flow-dependent nonlinear variations, thereby improving prediction at held-out higher-flow conditions.

\textbf{Qualitative comparison under an unseen inflow condition.}
Fig.~\ref{fig:10} compares the predicted pressure and velocity fields at the held-out inlet mass-flow rate of \(0.004~\mathrm{kg/s}\). Point-ResKAN reproduces the dominant spatial patterns of the CFD reference solution and maintains relatively small errors over most of the vascular domain. The largest discrepancies occur near the aneurysm neck and in regions of high curvature or steep velocity gradients, where the local flow structures are more difficult to approximate. Compared with Point-ResKAN, FNO and DeepONet exhibit more pronounced errors near bifurcations and geometrically complex regions. KA-PointNet also produces larger localized discrepancies, particularly in regions with rapidly varying flow. This behavior is consistent with the affine residual paths in Point-ResKAN improving feature propagation and the optimization of the Jacobi-based KAN mappings when the available inflow conditions are limited. Overall, Point-ResKAN more accurately preserves the principal pressure and velocity structures at an inflow condition excluded from training, demonstrating its potential as an efficient surrogate for inflow-conditioned hemodynamic field prediction.

\subsection{Results on transient velocity prediction}
\label{sec:transient}

The preceding experiments evaluate steady pressure--velocity prediction. We further assess Point-ResKAN on transient velocity prediction using pulsatile, non-Newtonian aneurysm-flow data. Following the formulation in Section~\ref{sec:transient_prediction}, the extended model predicts velocity increments from the current velocity state, spatial coordinates, a wall-boundary mask, and temporal encoding. This experiment evaluates the framework after task-specific training on transient CFD data.

\textbf{Dataset and Reference CFD Simulations.}
The reference data are obtained from the BenchAnXplore dataset.\cite{Lannelongue2026BenchAnXplore} The geometries are semi-idealized sidewall aneurysms formed by combining patient-derived aneurysm bulges from IntrA\cite{yang2020intra} with an idealized half-torus parent artery. The parent-vessel diameter is \(4~\mathrm{mm}\), and the aneurysm bulge diameters range from approximately \(3\) to \(9~\mathrm{mm}\). 

The reference simulations incorporate Carreau--Yasuda non-Newtonian rheology, a pulsatile parabolic inlet profile based on internal carotid artery measurements, a resistive outlet condition, and rigid no-slip walls. Following a \(0.2~\mathrm{s}\) linear ramp-up, two cardiac cycles are simulated. The first cycle is used for initialization, and the second is retained for analysis. The CFD time step is \(0.001~\mathrm{s}\), and velocity fields are stored at intervals of \(0.01~\mathrm{s}\).

\textbf{Remeshing and Interpolation Assessment.}
The dataset authors construct reduced-resolution representations using isotropic triangular surface meshes and unstructured tetrahedral volume meshes. A target element size of \(0.3~\mathrm{mm}\) yields fewer than \(25{,}000\) nodes and approximately \(120{,}000\) elements per case, without additional boundary-layer refinement. The source study examines target resolutions from \(0.15\) to \(0.4~\mathrm{mm}\) and reports round-trip interpolation errors of a few percent between the fine and coarse representations. This assessment characterizes errors introduced by transferring fields between meshes; it does not constitute an independent CFD mesh-convergence study.

\textbf{Temporal Sampling and Data Partition.}
We use \(50\) geometries, with \(48\) assigned to training and two held out for testing. Each geometry contains \(80\) stored frames over a cardiac cycle of \(0.8~\mathrm{s}\). Every second frame is retained, giving \(40\) frames with a prediction interval of \(0.02~\mathrm{s}\). This interval is distinct from the internal CFD time step. Each geometry contains approximately \(20{,}000\) spatial points. During training, \(5{,}000\) points are randomly sampled, whereas inference is evaluated on the complete available point clouds. One-step prediction uses the current CFD field as input. Autoregressive rollout starts from a CFD field and subsequently advances using the model's own predictions through the retained sequence.

\textbf{Training Configuration.}
The model is trained using Adam with a weight decay of \(0.05\) and a batch size of four geometries, with eight temporal samples selected per geometry. The learning rate is initially \(6\times10^{-4}\) and reduced to \(1\times10^{-5}\) at epoch \(100\). Training uses one-step transitions and the normalized MSE defined in Section~\ref{sec:transient_prediction}.

The binary wall mask has dimension \(d_B=1\), and the temporal encoding has dimension \(d_t=18\), giving a pointwise input dimension of \(6+d_B+d_t=25\). The three output components represent the predicted velocity increment. After adding this increment to the current state, zero velocity is imposed at wall nodes to enforce the stationary no-slip condition. This operation does not set the velocities at interior near-wall points to zero. 
The local and global mapping widths for PointNet are
\([64,128,256,256,256]\) and \([256,512,512]\), respectively.
KA-PointNet, PointNet(res-MLP), and Point-ResKAN use local and global mapping widths of
\([64,128,256]\) and \([256,512]\), respectively.
All models use a decoder with layer widths
\([768,512,256,128,3]\).
A common RMS velocity scale is computed exclusively from the training data, excluding wall nodes, and applied unchanged to the test data.

\textbf{One-Step and Autoregressive Prediction Results.}
Table~\ref{table:tran} compares the velocity prediction errors under both evaluation protocols. Point-ResKAN achieves the lowest reported errors among the evaluated models, with relative \(L_2\) errors of \(1.16\%\) for one-step prediction and \(6.89\%\) for autoregressive rollout. The corresponding RMSEs are \(2.86\) and \(14.68~\mathrm{mm/s}\), respectively. In comparison, KA-PointNet yields relative \(L_2\) errors of \(1.27\%\) and \(7.63\%\), while PointNet(res-MLP) yields \(1.56\%\) and \(8.53\%\).

\begin{figure}[t]
 \centering
 \includegraphics[width=0.9\linewidth]{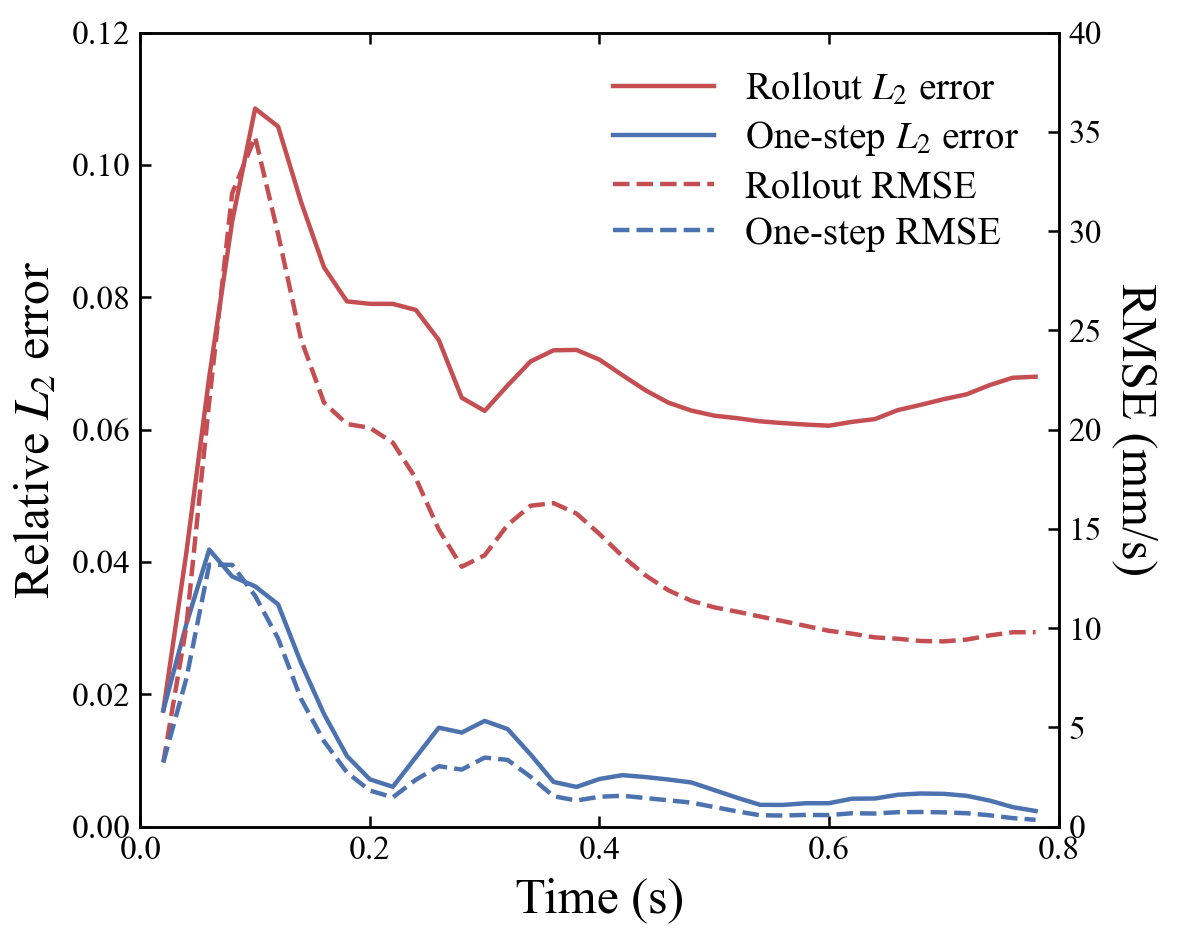}
 \caption{Temporal evolution of the relative \(L_2\) velocity error and RMSE over the evaluated cardiac cycle.}
 \label{fig:time_error}
\end{figure}

\begin{figure*}[htbp]
 \centering
 \includegraphics[width=\linewidth]{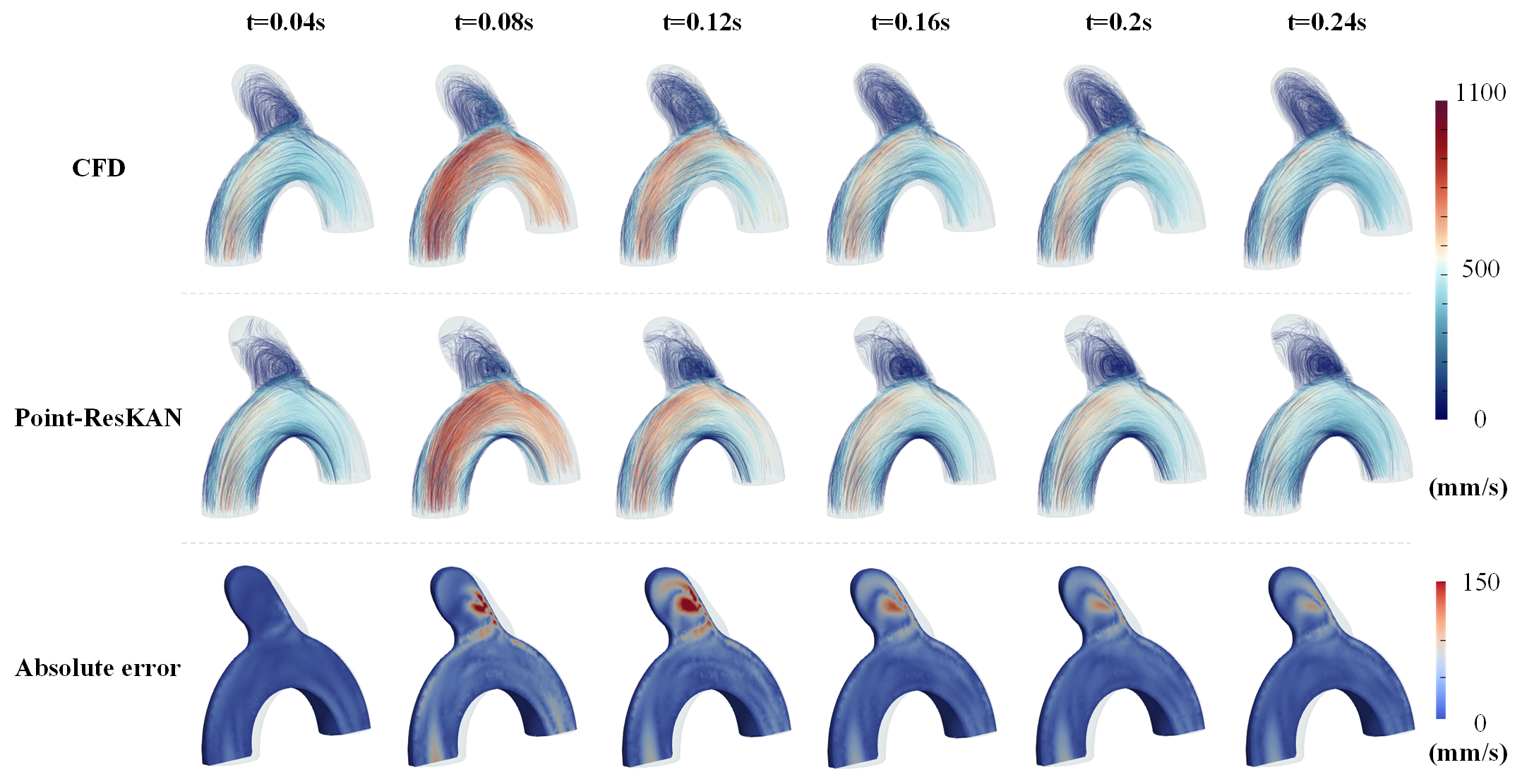}
 \caption{Predicted and CFD reference velocity fields during systole, together with the corresponding absolute-error distributions.}
 \label{fig:tr}
\end{figure*}

Rollout errors exceed one-step errors for every model, reflecting the additional difficulty of advancing from predicted rather than reference states. Fig.~\ref{fig:time_error} presents the error evolution through the cardiac cycle, and Fig.~\ref{fig:tr} compares the spatial velocity fields during systole. Point-ResKAN reproduces the principal flow patterns, although localized discrepancies remain near the aneurysm neck and dome. These results extend the evaluation to transient velocity fields under pulsatile, non-Newtonian conditions. They support the performance of the adapted framework on the two held-out geometries, while broader generalization to other inflow waveforms and sustained prediction over multiple cardiac cycles remains to be evaluated.

\begin{table}[t]
\centering
\footnotesize
\caption{Transient velocity prediction errors for autoregressive rollout and one-step prediction. Relative \(L_2\) errors are reported as percentages, and RMSE values are in \(\mathrm{mm/s}\).}
\label{table:tran}
\begin{tabular}{lcccc}
\hline\hline
\multirow{2}{*}{Model}
& \multicolumn{2}{c}{Autoregressive rollout}
& \multicolumn{2}{c}{One-step prediction} \\
\cmidrule(lr){2-3} \cmidrule(lr){4-5}
& Relative \(L_2\) (\%) & RMSE
& Relative \(L_2\) (\%) & RMSE \\
\hline
PointNet          & 8.21 & 16.96 & 1.37 & 3.33 \\
PointNet(res-MLP) & 8.53 & 18.34 & 1.56 & 3.82 \\
KA-PointNet       & 7.63 & 16.06 & 1.27 & 3.16 \\
Point-ResKAN      & \textbf{6.89} & \textbf{14.68}
                  & \textbf{1.16} & \textbf{2.86} \\
\hline\hline
\end{tabular}
\end{table}

\textbf{Benchmark Results and Architectural Comparisons.}
Table~\ref{table:tran} compares the four models under one-step prediction and autoregressive rollout. Point-ResKAN achieves the lowest reported errors under both protocols, with relative \(L_2\) errors of \(1.16\%\) and \(6.89\%\), respectively. These correspond to relative reductions of \(8.66\%\) and \(9.70\%\) compared with KA-PointNet. PointNet(res-MLP) does not improve upon PointNet in these experiments, whereas Point-ResKAN yields lower errors than KA-PointNet. This comparison supports the performance of the proposed ResKAN configuration in transient velocity prediction, although it does not isolate the contribution of the affine residual branch from other architectural differences.

Fig.~\ref{fig:time_error} shows higher rollout errors than one-step errors throughout the evaluated sequence. Rollout errors peak near \(t=0.1~\mathrm{s}\) and subsequently decrease with fluctuations, while one-step errors remain low during the latter part of the cycle. This difference is consistent with the propagation of prediction errors when model outputs are reused as inputs, whereas one-step evaluation starts from the CFD reference state at each update. The nonmonotonic rollout-error evolution also indicates that error magnitude varies with cardiac phase, rather than increasing uniformly with the number of prediction steps.

\textbf{Spatial Velocity-Field Comparisons.}
Fig.~\ref{fig:tr} compares the CFD reference fields and Point-ResKAN predictions at representative systolic instants, together with absolute-error maps. The predictions reproduce the main high-velocity regions along the curved parent vessel and the evolving velocity distribution within the aneurysm sac. Errors are smaller across much of the vessel and more pronounced near the aneurysm neck and dome, identifying regions where local prediction accuracy requires improvement. Together with the temporal error analysis, these comparisons support the model's ability to approximate the principal velocity-field features under the evaluated pulsatile, non-Newtonian conditions.

\subsection{Limitations}
\label{sec:limitations}

Despite its promising performance, Point-ResKAN has several limitations. First, the CFD data in the Aneumo dataset are generated under simplified physical assumptions, including steady inflow and rigid vessel walls. The additional transient benchmark incorporates pulsatile inflow, non-Newtonian blood rheology, and resistive outlet conditions, extending the evaluation to one-step and autoregressive velocity prediction.\cite{mousavi2024stentdeformation,gerdroodbary2026sacvolume} However, this evaluation remains limited to the prescribed inflow waveform and the selected dataset. Both hemodynamic datasets assume rigid vessel walls; consequently, vascular-wall deformation and fluid–structure interaction are not considered. Incorporating a wider range of inflow waveforms, compliant-wall models, and more physiologically realistic outlet conditions will be necessary to improve the fidelity of future hemodynamic predictions.

Second, the medical validation is currently limited to intracranial aneurysm geometries. Although the Ahmed-body experiment demonstrates the ResKAN architecture's transferability to a distinct geometry-conditioned flow problem, it does not establish clinical generalization across vascular anatomies. Coronary arteries, carotid bifurcations, and the abdominal aorta may exhibit substantially different geometric characteristics, flow regimes, and sensitivities to boundary conditions. Meanwhile, using patient-specific models requires additional uncertainty assessment, including the effects of uncertain vascular geometry and boundary conditions, along with independent validation. Evaluation on larger and more diverse vascular datasets, including patient-specific measurements and prospective clinical data, is therefore required before the proposed surrogate can be considered broadly applicable to hemodynamic prediction.

\section{Conclusions}
We presented Point-ResKAN, a neural-operator-inspired point-cloud surrogate for predicting pressure–velocity fields from vascular geometry and prescribed inflow conditions. The framework combines dual-branch PointNet-style encoding with residual Jacobi-based KAN layers, enabling pointwise geometric information, global vascular context, and inflow features to be integrated within a nonlinear latent representation. The affine residual branches provide direct paths for feature and gradient propagation while retaining the nonlinear approximation capability of the Jacobi-based KAN mappings.

Experiments on the intracranial-aneurysm dataset show that Point-ResKAN consistently achieves lower pressure and velocity errors than the considered baseline methods. It also maintains improved predictive accuracy when evaluated on vascular geometries and inflow conditions not included in training. The empirical convergence study and controlled comparisons with KA-PointNet further support the contribution of the affine residual construction to training stability and generalization. Additional experiments on transient CFD data incorporating pulsatile inflow, non-Newtonian blood rheology, and resistive outlet conditions extend the evaluation to one-step and autoregressive velocity prediction. Point-ResKAN achieves lower errors than the compared models in both settings under the evaluated conditions. In addition, the Ahmed-body experiments demonstrate that the proposed ResKAN architecture can be extended to geometry- and flow-conditioned surface-pressure prediction beyond vascular hemodynamics.

Once trained, Point-ResKAN provides pressure–velocity predictions in substantially less time than the corresponding CFD simulations, supporting its use as an efficient surrogate for rapid hemodynamic evaluation. Nevertheless, further work is required to evaluate a wider range of pulsatile inflow waveforms and patient-specific outlet conditions, incorporate compliant vascular walls, and assess generalization to independent patient cohorts through broader clinical validation. Overall, Point-ResKAN provides an accurate and computationally efficient approach to point-cloud-based flow-field prediction and offers a flexible architecture for surrogate modeling of parameterized three-dimensional fluid flows.

\begin{acknowledgments}
This research is supported by the National Natural Science
Foundation of China NSFC 12071345.
\end{acknowledgments}
\vspace{-\baselineskip}
\section*{Author Declarations}
\subsection*{Conflicts of Interest}
The authors have no conflict of interest.
\subsection*{Data Availability}
The Python code is publicly available on the following GitHub repository at \url{https://github.com/pridgetulin/Point-ResKAN}.

\subsection*{Author Contributions}
Jiaheng Kang: Methodology, investigation, visualization, and writing--original draft. Jun Liu: Methodology and writing--review and editing. Yuping Duan: Conceptualization, methodology, supervision, and writing--review and editing. All authors have read and approved the final version of the manuscript.

\appendix*
\section{Additional Evaluation Results on the Ahmed-Body Benchmark}
\label{sec:ahmed_ablation}

We use the Ahmed-body benchmark~\cite{li2023geometry} as a supplementary architecture-level evaluation of the ResKAN feature-transformation module. The task is to predict surface-pressure distributions from surface geometry and scalar flow parameters. Models are trained specifically for this task, without the dedicated inlet-point-cloud encoder used in the aneurysm experiments. Comparisons of network variants, Jacobi basis parameters, and polynomial degree characterize the module's performance within this external-aerodynamics setting. The benchmark complements the aneurysm experiments but does not establish cross-domain transfer, a complete ablation of the aneurysm-prediction architecture, or clinical generalization.

\begin{figure*}
 \centering
 \includegraphics[width=0.8\linewidth]{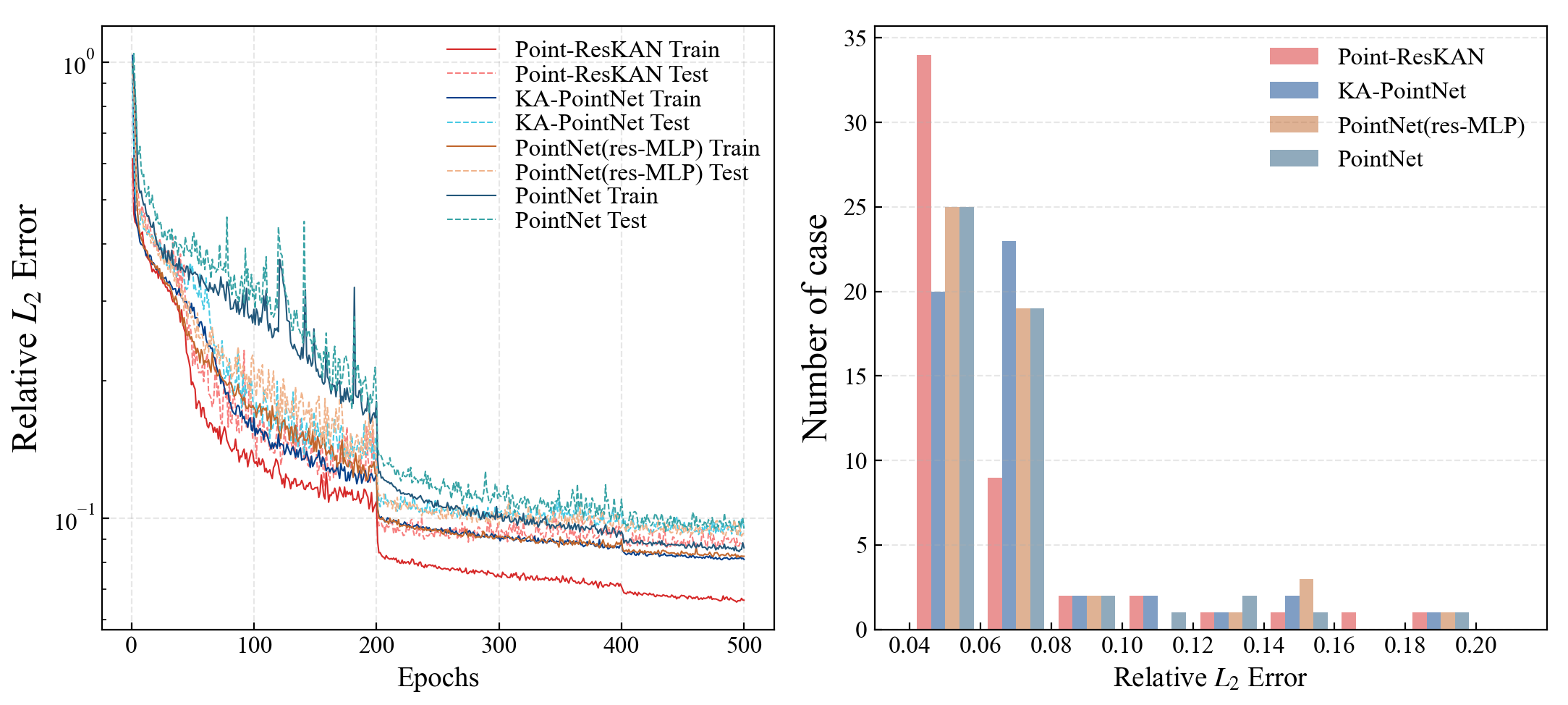}
 \caption{Comparison of the loss curve and relative $L_2$ error distribution for the different surrogate models in the Ahmed dataset.}
 \label{fig:11}
\end{figure*}

\textbf{Ahmed-Body dataset.} The reference steady-flow solutions are generated using a GPU-accelerated CFD solver for the three-dimensional Navier--Stokes equations. The freestream velocity \(U_{\infty}\) ranges from \(10\) to \(70~\mathrm{m/s}\), with the corresponding Reynolds number satisfying $\operatorname{Re}\in\left[4.35\times10^{5},6.82\times10^{6}\right]$. The dataset contains \(500\) training cases and \(51\) test cases. For each case, Point-ResKAN takes the sampled surface geometry together with the flow parameters \(U_{\infty}\) and \(\operatorname{Re}\) as inputs and predicts the pressure values at the sampled points on \(\Gamma_{\boldsymbol{\eta}}\). This benchmark is used to compare the effects of the affine residual branch introduced in ResKAN.

\begin{figure*}
 \centering
 \includegraphics[width=0.89\linewidth]{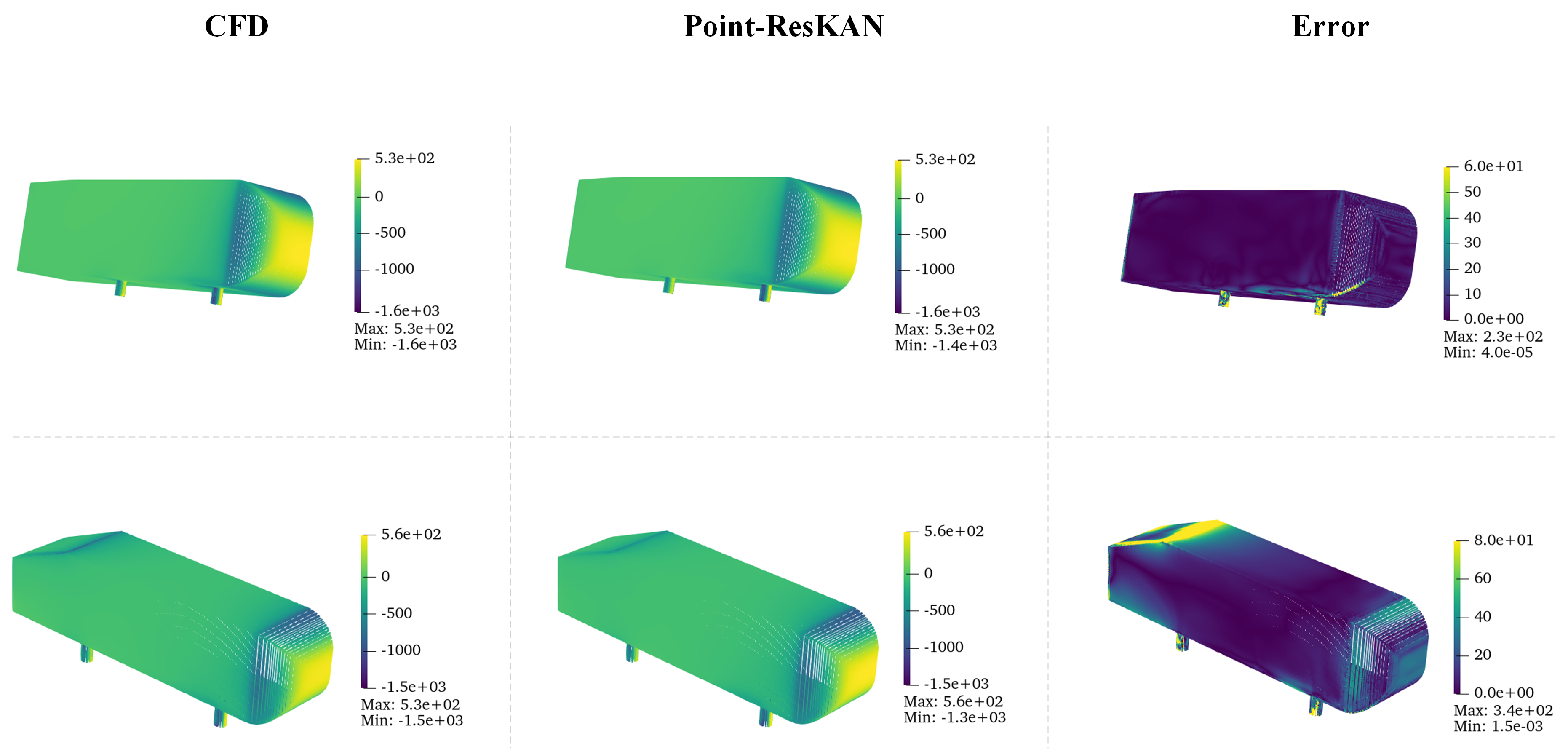}
 \caption{ Learning curves and case-wise test relative \(L_2\) error distributions for the compared point-based models on the Ahmed-body benchmark.}
 \label{fig:12}
\end{figure*}

\textbf{Training configuration.} For the Ahmed-body dataset, all models are trained using the Adam optimizer with a weight decay of 0.05 and a batch size of 8. The initial learning rate is set to $6\times10^{-4}$ and reduced to $1\times10^{-4}$ and $5\times10^{-5}$ at epochs 200 and 400, respectively. During each training epoch, 20,000 points are randomly sampled from each geometry. The models are optimized using the relative $L_2$ loss scaled by a factor of 100. For a fair comparison, the local and global mapping dimensions of PointNet are set to $[5, 64, 128, 256, 256, 256, 256]$ and $[256, 256, 512, 1024]$, respectively. KA-PointNet, PointNet with residual multilayer perceptron (res-MLP) and Point-ResKAN share the same local and global mapping dimensions of $[5, 32, 64, 128, 256]$ and $[256, 1024]$, respectively. In Point-ResKAN, the edge functions are parameterized using second-order Jacobi polynomials with $\alpha=\beta=1.0$. All models employ an identical decoder with layer dimensions $[1280, 512, 256, 128, 1]$. Normalize each case individually. The training error is the normalized error, and the test error is the denormalized error.

\textbf{Convergence behavior and error distribution.}
The left of Fig.~\ref{fig:11} compares the convergence trajectories of Point-ResKAN, KA-PointNet, PointNet, and PointNet(res-MLP). Point-ResKAN converges to a lower training error than the two point-based baselines and exhibits smaller oscillations during training. The case-wise error distribution in the right of Fig.~\ref{fig:11} shows that the relative \(L_2\) errors of Point-ResKAN for most test geometries fall within the interval \([0.04,0.06]\). The comparatively concentrated distribution indicates more consistent surface-pressure predictions across geometries not included in training. Together, these results are consistent with the affine residual paths improving the optimization of the Jacobi-based KAN layers and reducing variation in prediction accuracy across different Ahmed-body geometries.

\textbf{Sensitivity to Jacobi basis parameters and polynomial degree.} We examine four Jacobi basis settings, \(\alpha=\beta\in\{-0.5,0,0.5,1\}\), with polynomial degree \(n=2\). As shown in Table~\ref{table:j}, the corresponding test relative \(L_2\) errors are \(6.79\%\), \(6.69\%\), \(6.64\%\), and \(6.51\%\), respectively. Thus, \(\alpha=\beta=1\) achieves the lowest error among the tested settings. Increasing the polynomial degree to \(n=3\) yields a test error of 6.54\% and increases the training time from \(9.8\) to \(12.5~\mathrm{s}\) per epoch. These results describe sensitivity to the selected polynomial configurations on the Ahmed-body benchmark; they do not establish an optimal configuration across other datasets or architectures.

\begin{table}[htbp]
\centering
\caption{Training and test relative \(L_2\) errors for different Jacobi basis parameters on the Ahmed-body benchmark (\(n=2\)).}
\label{table:j}
\begin{tabular}{l S[table-format=2.2] S[table-format=2.2]}
\hline\hline
\text{Parameters} & \text{Train error (\%)} & \text{Test error (\%)} \\
\hline
$\alpha=\beta=-0.5$& 7.03 & 6.79 \\
$\alpha=\beta=0$   & 6.89 & 6.69 \\
$\alpha=\beta=0.5$ & 6.72 & 6.64 \\
$\alpha=\beta=1$   & 6.6 & 6.51  \\
\hline\hline
\end{tabular}
\end{table}

\textbf{Benchmark comparison and ablation results.}
Table~\ref{table:a} summarizes the quantitative comparison on the Ahmed-body benchmark. Point-ResKAN achieves relative \(L_2\) errors of \(6.60\%\) and \(6.51\%\) on the training and test sets, respectively. Its test error is lower than those of the neural-operator baselines GINO (\(8.31\%\)) and FNO (\(12.59\%\)), demonstrating improved surface-pressure prediction on geometries not included in training. The controlled ablation comparison further shows that Point-ResKAN outperforms KA-PointNet (\(7.06\%\)), PointNet (\(8.09\%\)), and PointNet(res-MLP) (\(7.32\%\)) on the test dataset. Because Point-ResKAN and KA-PointNet use matched Jacobi-polynomial settings and latent feature dimensions, their performance difference provides direct evidence of the contribution of the affine residual branches. These results indicate that ResKAN improves both prediction accuracy and generalization in geometry-conditioned surface-pressure modeling.

\begin{table}[htbp]
\centering
\caption{Comparison of the relative $L_2$ error by different models.}
\label{table:a}
\begin{tabular}{l S[table-format=2.2] S[table-format=2.2]}
\hline\hline
\text{Model} & ~~\text{Training error (\%)} ~~ & \text{Test error (\%)} \\
\hline
MeshGraphNet   & 9.08  & 13.88 \\
UNet(interp)   & 9.93  & 11.16 \\
FNO(interp)    & 12.97 & 12.59 \\
GINO           & 9.34  & 8.31  \\
PCNO           & {--}  & 6.82  \\
PointNet       & 9.97  & 8.09  \\
PointNet(res-MLP) & 8.3  & 7.32  \\
KA-PointNet    & 8.12  & 7.06  \\
\hline
Point-ResKAN   & \textbf{6.6}   & \textbf{6.51}  \\
\hline\hline
\end{tabular}
\end{table}

\textbf{Qualitative surface-pressure prediction.}
Fig.~\ref{fig:12} presents representative test cases with relatively low and high prediction errors. For cases with comparatively smooth surface-pressure variations, Point-ResKAN closely reproduces the CFD reference distribution. In more challenging geometries, the largest discrepancies are concentrated near the rear-slant and sharp trailing-edge regions, where flow separation and adverse pressure gradients produce rapid spatial variations in surface pressure. Despite these localized errors, Point-ResKAN preserves the dominant pressure patterns and provides closer agreement with the reference solutions than the compared baselines. These results further indicate that the ResKAN module improves geometry-conditioned pressure prediction while retaining sensitivity to localized aerodynamic features.

\textbf{Training Cost and Repeated-Run Results.} The average training time per epoch is \(7.1~\mathrm{s}\) for PointNet, \(7.4~\mathrm{s}\) for PointNet(res-MLP), \(8.7~\mathrm{s}\) for KA-PointNet, and \(9.8~\mathrm{s}\) for Point-ResKAN. Point-ResKAN achieves lower reported test errors than these comparators with additional training cost. Across five independent random seeds, its test relative \(L_2\) error is \(6.542\% \pm 0.155\%\) (mean \(\pm\) standard deviation). This statistic characterizes the training variability of Point-ResKAN on this benchmark; it does not replace repeated-run comparisons across competing models.

The repeated-run results characterize training variability on the Ahmed-body benchmark, while the hemodynamic experiments assess deterministic prediction accuracy. Neither provides calibrated uncertainty estimates for individual hemodynamic predictions. Extending the framework to quantify model uncertainty and propagate uncertainty in vascular geometry and boundary measurements is an important direction for further evaluation. Such estimates would require calibration and coverage assessment, together with independent validation, before use in patient-specific decision support.

For the Ahmed-body cases considered here, each reference CFD simulation requires approximately \(7\) to \(19\) hours, whereas the trained Point-ResKAN model predicts the surface-pressure values at all sampled points in less than \(1\) second. Excluding the one-time training cost, this reduction in inference time enables rapid evaluation of steady-state pressure fields across different geometries and inflow conditions.

\section*{Reference}
\nocite{*}
\bibliography{ref}% Produces the bibliography via BibTeX.

\end{document}